\documentclass[conference,a4paper]{APSIPA2021}
\usepackage{multirow}
\usepackage[dvips]{graphicx}
\usepackage{amsmath}
\usepackage[psamsfonts]{amssymb}
\usepackage{amsxtra}
\usepackage{threeparttable}
\usepackage{bm}
\usepackage{cite}
\usepackage{url}
\let\labelindent\relax
\usepackage{enumitem}
\usepackage{subcaption}
\usepackage{caption}

\begin{document}

\title{Spatially varying white balancing for mixed and non-uniform illuminants}

\author{%
\authorblockN{%
Teruaki Akazawa\authorrefmark{1},
Yuma Kinoshita\authorrefmark{1} and
Hitoshi Kiya\authorrefmark{1}
}
\authorblockA{%
\authorrefmark{1}
Tokyo Metropolitan University, Tokyo, Japan\\
E-mail: akazawa-teruaki@ed.tmu.ac.jp, ykinoshita@tmu.ac.jp, kiya@tmu.ac.jp Tel/Fax: +81-42-585-8454}
}

\maketitle
\thispagestyle{empty}

\begin{abstract}
In this paper, we propose a novel white balance adjustment, called ``spatially varying white balancing,'' for single, mixed, and non-uniform illuminants.
By using n diagonal matrices along with a weight, the proposed method can reduce lighting effects on all spatially varying colors in an image under such illumination conditions.
In contrast, conventional white balance adjustments do not consider the correcting of all colors except under a single illuminant.
Also, multi-color balance adjustments can map multiple colors into corresponding ground truth colors, although they may cause the rank deficiency problem to occur as a non-diagonal matrix is used, unlike white balancing.
In an experiment, the effectiveness of the proposed method is shown under mixed and non-uniform illuminants, compared with conventional white and multi-color balancing.
Moreover, under a single illuminant, the proposed method has almost the same performance as the conventional white balancing.
\end{abstract}

\section{Introduction}
Image segmentation and object recognition are required to decompose an image into meaningful regions.
A typical approach to this problem is to assign a single class to each pixel in an image.
However, such hard segmentation is far from ideal when the distinction between meaningful regions is ambiguous, such as in the cases of objects with motion blur or color distortion due to illumination and image enhancement\cite{Color_Constancy_effect_on_DNN_2019_ICCV,Fast_Soft_Segmentation_2020_CVPR,kinoshita-san-Hue,kinoshita-san-APSIPA-HUE,kinoshita-san-Semantic_2019}.
Accordingly, we aim to reduce various illumination effects on all colors in an image.

A change in illumination affects the pixel value of an image taken with an RGB digital camera.
White balancing is a technique that reduces color distortion due to illumination changes (i.e., lighting effects).
By applying white balancing, color constancy correction, that is, to make all colors in an image constant, can be maintained regardless of the illumination.
In white balancing, source white with remaining lighting effects and corresponding ground truth white are used for designing a matrix that maps source white into ground truth white. 
However, because only a single illuminant is assumed in the color mapping of white balancing, spatially varying colors caused by mixed or non-uniform illuminants cannot be adjusted by using white balancing.

Multi-color balance adjustments have also been proposed to improve the color constancy correction performance of conventional white balancing\cite{ChengsBeyondWhite,Akazawa_2021_Lifetech_Multi-color,akazawa_ICIP_n-color_2021}.
In these methods, multiple colors including both achromatic and chromatic colors can be used for designing a matrix that maps multiple colors into ground truth colors.
However, deciding the number or the combination of multiple colors and estimating multiple colors are difficult.

Accordingly, in this paper, we propose a novel white balancing called ``spatially varying white balancing'' for adjusting spatially varying colors caused by mixed or non-uniform illuminants.
The proposed method is performed by using n diagonal matrices designed from each spatially varying white point.
When designing one matrix assuming a single illuminant, conventional white balancing cannot reduce lighting effects on such colors.
Also, multi-color balancing can be applied to this scenario by regarding spatially varying white as multiple colors.
However, in multi-color balancing, the selection of various white points may cause the rank deficiency problem to occur because the colors are of the same type.
In contrast, because all the matrices are diagonal in the proposed method, this problem does not occur.

In an experiment, we used several images taken under a single illuminant and mixed and non-uniform illuminants.
The proposed method was confirmed to adjust colors as well as white balancing under a single illuminant and outperformed the conventional methods under mixed and non-uniform illuminants.

\section{Related work}\label{sec:related_work}

We summarize conventional methods for color constancy and problems with these methods.

\subsection{White balance adjustment}

By using white balancing, lighting effects on a white region in an image are accurately corrected if the white region under illumination is correctly estimated.

White balancing is performed by
\begin{equation}
\label{eqn:chromadaptWB}
\bm{P}_{\rm{WB}} = \bf{{M}_{\rm{WB}}} \bm{{P}}_{\rm{XYZ}}
,
\end{equation}
where ${\bm{P}_{\rm{XYZ}}=(X_{\rm{P}},Y_{\rm{P}},Z_{\rm{P}})^\top}$ is a pixel value of an image in the XYZ color space\cite{CIE_XYZ_color_space}, and ${\bm{P}_{\rm{WB}}=(X_{\rm{WB}},Y_{\rm{WB}},Z_{\rm{WB}})^\top}$ is that of a white balanced image\cite{brucelindbloom}.
$\bf{M}_{\rm{WB}}$ in (\ref{eqn:chromadaptWB}) is given as
\begin{equation} 
\label{eqn:Mwb}
    \bf{M}_{\rm{WB}} = {\bf{M}_{\rm{A}}}^{\rm{-1}}
    \left(
    \begin{array}{cccc}
    {\rho_{\rm{D}}}/{\rho_{\rm{S}}} & 0 & 0 \\
   0 & {\gamma_{\rm{D}}}/{\gamma_{\rm{S}}} & 0 \\
   0 & 0 & {\beta_{\rm{D}}}/{\beta_{\rm{S}}} \\
    \end{array}
    \right)
    \bf{M}_{\rm{A}}
.
\end{equation}
$\bf{M}_{\rm{A}}$ with a size of 3$\times$3 is decided in accordance with an assumed chromatic adaptation transform\cite{brucelindbloom}.
$(\rho_{\rm{S}},\gamma_{\rm{S}},\beta_{\rm{S}})^\top$ and $(\rho_{\rm{D}},\gamma_{\rm{D}},\beta_{\rm{D}})^\top$ are calculated from a source white point $(X_{\rm{S}},Y_{\rm{S}},Z_{\rm{S}})^\top$ in an input image and a ground truth white point $(X_{\rm{D}},Y_{\rm{D}},Z_{\rm{D}})^\top$ as
\begin{equation}
\label{eqn:color-xfer}
\left(\!
\begin{array}{cccc}
   {\rho_{\rm{S}}} \\
   {\gamma_{\rm{S}}} \\
   {\beta_{\rm{S}}} \\
  \end{array} 
\!  \right)
  = \bf{M}_{\rm{A}} \left(\!
  \begin{array}{cccc}
   {X_{\rm{S}}} \\
   {Y_{\rm{S}}} \\
   {Z_{\rm{S}}} \\
  \end{array}
\! \right) \: {\rm{,and}} \:
 \left(\!
\begin{array}{cccc}
   {\rho_{\rm{D}}} \\
   {\gamma_{\rm{D}}} \\
   {\beta_{\rm{D}}} \\
  \end{array} 
\!  \right)
  = \bf{M}_{\rm{A}} \left(\!
  \begin{array}{cccc}
   {X_{\rm{D}}} \\
   {Y_{\rm{D}}} \\
   {Z_{\rm{D}}} \\
  \end{array}
\!    \right) 
.
\end{equation}
Using the 3$\times$3-identity matrix as $\bf{M}_{\rm{A}}$ indicates that white balancing is performed in the XYZ color space.
Otherwise, von Kries's\cite{vonKriesOriginal} and Bradford's\cite{BradfordOriginal} chromatic adaptation transforms were also proposed for reducing lighting effects on colors other than white under the framework of white balancing.
For example, under the use of Bradford's model, $\bf{M}_{\rm{A}}$ is given as
\begin{equation}  
\label{eqn:Ma} 
  \bf{M}_{\rm{A}} = \left(
  \begin{array}{cccc}
   0.8951 & 0.2664 & -0.1614 \\
   -0.7502 & 1.7135 & 0.0367 \\
   0.0389 & -0.0685 & 1.0296 \\
  \end{array}
    \right)
.
\end{equation}

White balancing is a technique that maps a source white point under a single illuminant into a ground truth one as in (\ref{eqn:Mwb}).
However, the conventional white balancing does not consider the adjusting of spatially varying colors caused under mixed or non-uniform illuminants.
Accordingly, it suffers from such illumination in terms of color constancy.

\subsection{Multi-color balancing}
Various chromatic adaptation transforms such as those of von Kries\cite{vonKriesOriginal}, Bradford\cite{BradfordOriginal}, and the latest CAM16\cite{CAM16original} were proposed and consider both chromatic and achromatic colors under the framework of white balancing.
In contrast, multi-color balance adjustments\cite{ChengsBeyondWhite,Akazawa_2021_Lifetech_Multi-color} were proposed to further relax the limitation that white balancing has.
In these methods, multiple colors including chromatic colors are used for designing a matrix that maps the multiple colors into ground truth colors.
However, multi-color balance adjustments have three problems: selecting the number of colors, selecting colors, and estimating chromatic colors automatically.
The proposed method is a type of white balancing, so it can estimate white regions by using an estimation method\cite{Max_RGB,Grey_World_Theory,Grey_Edge,Cheng_PCA,APAP_bias_illumination_estimation,AWB_goes_wrong,Deep_WB_Editing} for white balancing in addition to having no need for colors to be selected.
\section{Proposed method}
Here, our novel white balancing method, called ``spatially varying white balancing,'' is proposed.
\subsection{Scenario}\label{sec:proposed_scenario}
\begin{figure}[tb]
\captionsetup[subfigure]{justification=centering}
\begin{minipage}[b]{0.495\linewidth}
  \centering
  \centerline{\includegraphics[keepaspectratio, scale=0.21]{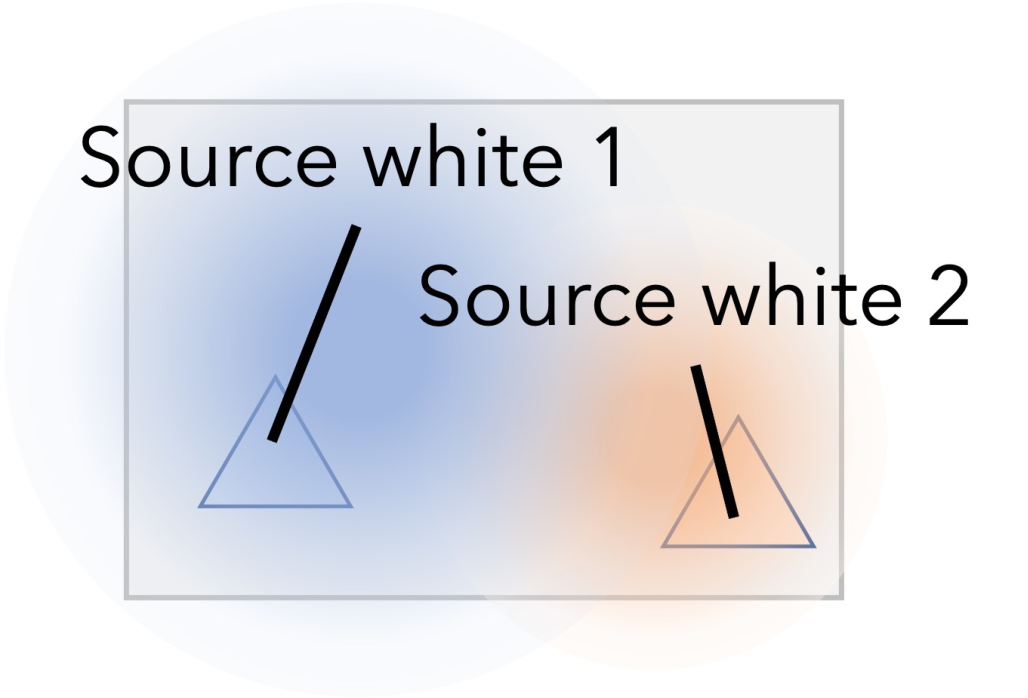}}
 \subcaption{}\label{fig:assumed_scenario_mixed}\medskip
\end{minipage}
\begin{minipage}[b]{0.495\linewidth}
  \centering
  \centerline{\includegraphics[keepaspectratio, scale=0.205]{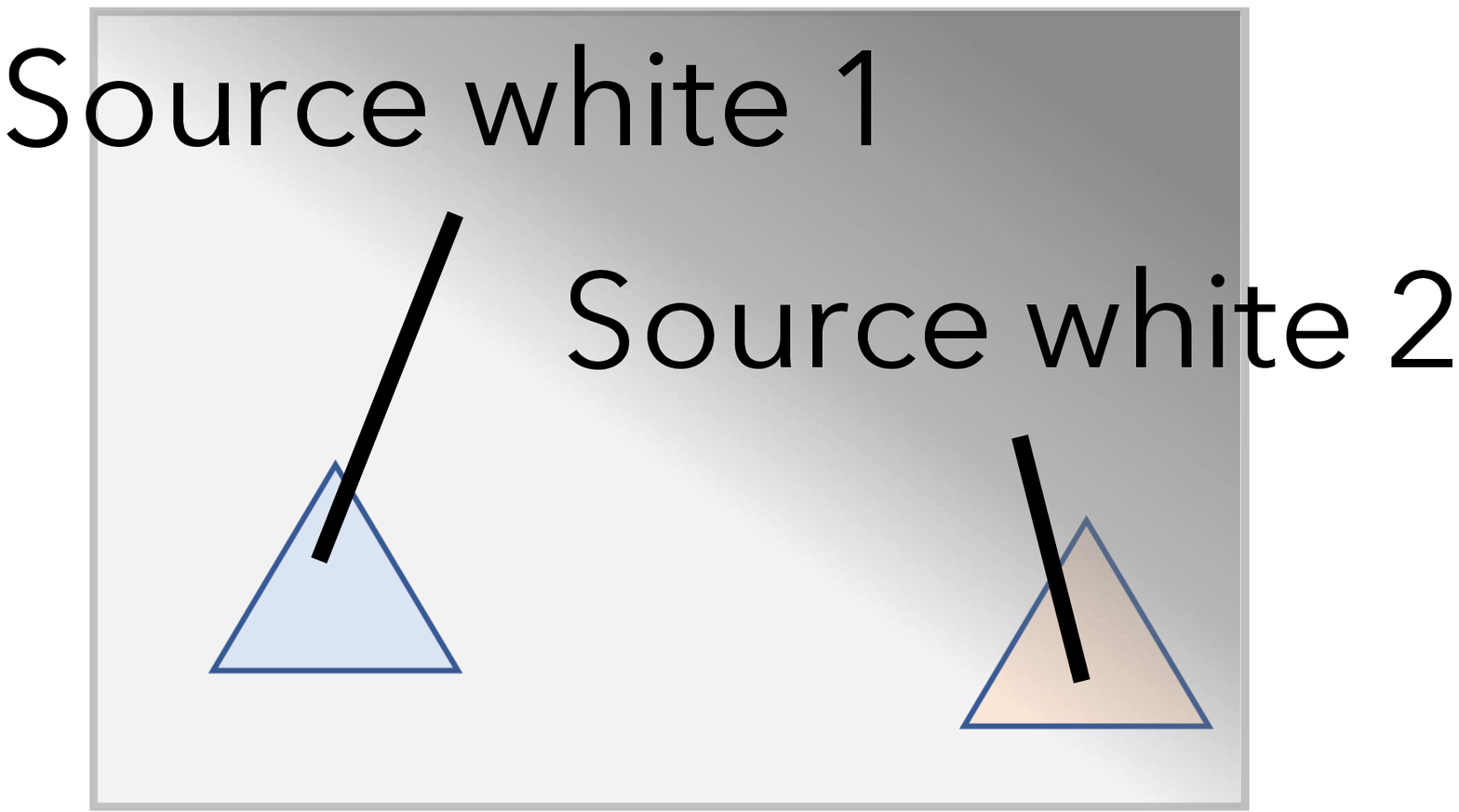}}
 \vspace{0.415cm}
 \subcaption{}\label{fig:assumed_scenario_non-uniform}\medskip
\end{minipage}
%
\caption{Overview of scenario.
(\subref{fig:assumed_scenario_mixed}) Mixed-illuminant and (\subref{fig:assumed_scenario_non-uniform}) non-uniform illuminant.}
\label{fig:assumed_scenario}
\end{figure}

Conventional white balancing assumes a single illuminant in general.
In contrast, in this paper, various lighting conditions such as mixed and non-uniform illuminants are considered in addition to a single illuminant.
Under such conditions, each white region in an image may have a different color as shown in Fig. \ref{fig:assumed_scenario}, so conventional white balancing cannot correct spatially varying white regions and other colors.
%
%
By individually adjusting these white regions, the proposed method aims to achieve color constancy correction for spatially varying colors caused by mixed and non-uniform illuminants.
\subsection{Spatially varying white balancing}
The proposed method is carried out by using n matrices, and the matrices are combined with weights; in comparison, the conventional methods including multi-color balancing are carried out with a single matrix as in (\ref{eqn:Mwb}). 
A pixel ${\bm{P}'_{\rm{WB}}}$ color balanced by the proposed method is given by
\begin{equation}
\label{eqn:chromadaptMLT}
\bm{P}'_{\rm{WB}} = {\bf{M}'_{\rm{WB}}} {\bm{P}}_{\rm{XYZ}}
,
\end{equation}
where
\begin{equation}
\label{eqn:M_MLT}
{\bf{{M}'_{\rm{WB}}}} = k_{1} {\bf{M}}_{1} + k_{2} {\bf{M}}_{2} + \cdots + k_{n} {\bf{M}}_{n}.
\end{equation}
$n$ is the number of spatially varying source white points that we focus on, and $k_{m}$, $m \in \{1,2,\cdots,n\}$, is a weight of ${\bf{M}}_{m}$. 
In (\ref{eqn:M_MLT}), ${\bf{M}}_{m}$ is designed in a similar way to white balancing. 
Let ${\bm{S}}_{m}=(X_{{\rm{S}}m},Y_{{\rm{S}}m},Z_{{\rm{S}}m})^\top$ and ${\bm{G}}_{m}=(X_{{\rm{G}}m},Y_{{\rm{G}}m},Z_{{\rm{G}}m})^\top$ be a source white point and a ground truth one in the XYZ color space, respectively. ${\bf{M}}_{m}$ is then given as
\begin{equation} 
\label{eqn:Mm}
    {\bf{M}}_{m} = {\bf{M}_{\rm{A}}}^{\rm{-1}}
    \left(
    \begin{array}{cccc}
    {\rho'_{\rm{D}}}/{\rho'_{\rm{S}}} & 0 & 0 \\
   0 & {\gamma'_{\rm{D}}}/{\gamma'_{\rm{S}}} & 0 \\
   0 & 0 & {\beta'_{\rm{D}}}/{\beta'_{\rm{S}}} \\
    \end{array}
    \right)
    \bf{M}_{\rm{A}}
,
\end{equation}
where
\begin{equation}
\label{eqn:color-xfer_Mm}
\left(\!\!\!
\begin{array}{cccc}
   {\rho'_{\rm{S}}} \\
   {\gamma'_{\rm{S}}} \\
   {\beta'_{\rm{S}}} \\
  \end{array} 
\!\!\!  \right)
 \!\! = \! \bf{M}_{\rm{A}} \left(\!\!\!
  \begin{array}{cccc}
   {X_{{\rm{S}}m}} \\
   {Y_{{\rm{S}}m}} \\
   {Z_{{\rm{S}}m}} \\
  \end{array}
\!\!\!\!  \right) \: {\rm{, and}} \:
 \left(\!\!\!
\begin{array}{cccc}
   {\rho'_{\rm{D}}} \\
   {\gamma'_{\rm{D}}} \\
   {\beta'_{\rm{D}}} \\
  \end{array} 
\!\!\!  \right)
\!\!\!  = \! \bf{M}_{\rm{A}} \left(\!\!\!
  \begin{array}{cccc}
   {X_{{\rm{G}}m}} \\
   {Y_{{\rm{G}}m}} \\
   {Z_{{\rm{G}}m}} \\
  \end{array}
\!\!\!\!  \right) 
.
\end{equation}
In a manner like conventional white balancing, the 3$\times$3-identity matrix is used as $\bf{M}_{\rm{A}}$ when the proposed method is performed in the XYZ color space.
Other models such as those by von Kries and Bradford can also be used as $\bf{M}_{\rm{A}}$.

When the spatial pixel coordinate of ${\bm{P}}_{\rm{XYZ}}$ is closer to that of source white ${\bm{S}}_{m}$ than the other source white, ${\bf{M}}_{m}$ designed with ${\bm{S}}_{m}$ should more contribute to adjusting ${\bm{P}}_{\rm{XYZ}}$ than the rest of the matrices.
Hence, to measure the distance between the coordinates of ${\bm{P}}_{\rm{XYZ}}$ and ${\bm{S}}_{m}$, the Euclidean distance is calculated as
\begin{equation}
\label{eqn:XYZ_pixel_distance}
d_{m} = \sqrt{ \left( x_{{\rm{S}}m} - x_{\rm{P}} \right)^2 +  \left( y_{{\rm{S}}m} - y_{\rm{P}} \right)^2 },
\end{equation}
where $ \left( x_{{\rm{S}}m} , y_{{\rm{S}}m} \right)$ is a pair of coordinates of ${\bm{S}}_{m}$, and $ \left( x_{\rm{P}} , y_{\rm{P}} \right)$ is those of ${\bm{P}}_{\rm{XYZ}}$.
A smaller $d_{m}$ means that $ \left( x_{\rm{P}} , y_{\rm{P}} \right)$ is closer to $ \left( x_{{\rm{S}}m} , y_{{\rm{S}}m} \right)$. 
Because $k_{m}$ in (\ref{eqn:M_MLT}) should be larger under a smaller $d_{m}$, the inverse proportion to $d_{m}$ is calculated as
\begin{equation}
d'_{m} = \frac{1}{d_{m}}
.
\end{equation}
To reduce the total value of weights to 1 in (\ref{eqn:M_MLT}), $k_{m}$ is given as
\begin{equation}
\label{eqn:coefficient_k}
k_{m} = \frac{d'_{m}}{d'_{1}+d'_{2}+\cdots+d'_{n}}
.
\end{equation}
Note that, in (\ref{eqn:coefficient_k}), $k_{m}$ will be infinite if input pixel ${\bm{P}}_{\rm{XYZ}}$ is equal to ${\bm{S}}_{m}$ (i.e., $d_{m}=0$). 
In this case, let $k_{m}$ be a value of 1, and let the other weights be a value of zero.

In our scenario, since spatially varying white regions have different colors respectively, conventional multi-color balancing can also be applied by designing a matrix from the varying white. However, selecting multiple white points for the matrix design may cause the rank deficiency problem to occur for a single matrix.
In contrast, in the proposed method, multiple diagonal matrices given by each different white point are used so that such a problem is never caused.
\section{Experiment}
We conducted an experiment to confirm the effectiveness of the proposed method.
\subsection{Experiment setup}
\begin{figure}[tb]
\captionsetup[subfigure]{justification=centering}
\begin{minipage}[b]{0.32\linewidth}
  \centering
  \centerline{\includegraphics[keepaspectratio, scale=0.1]{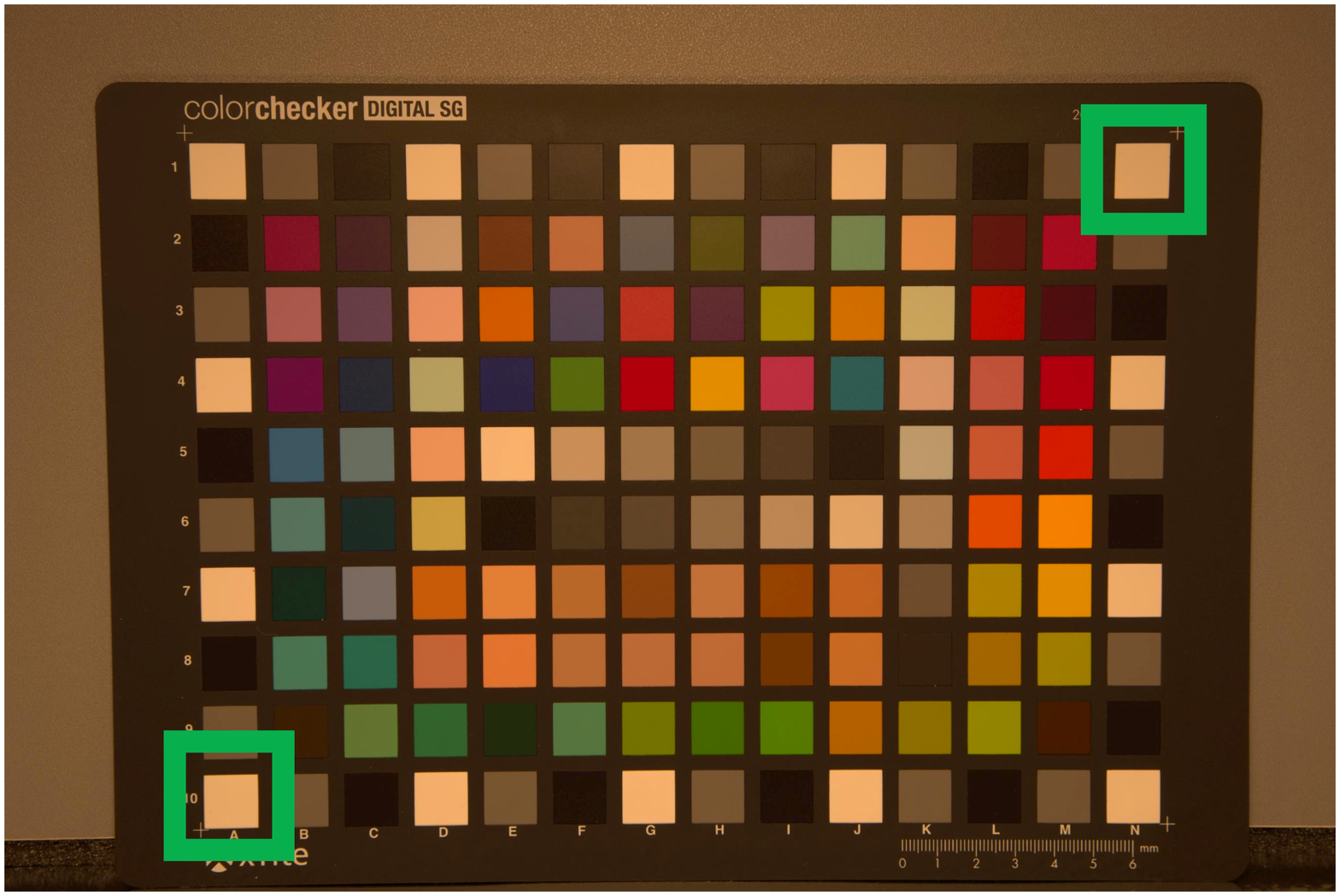}}
 \subcaption{}\label{fig:single-inside_input}\medskip
\end{minipage}
\begin{minipage}[b]{0.32\linewidth}
  \centering
  \centerline{\includegraphics[keepaspectratio, scale=0.103]{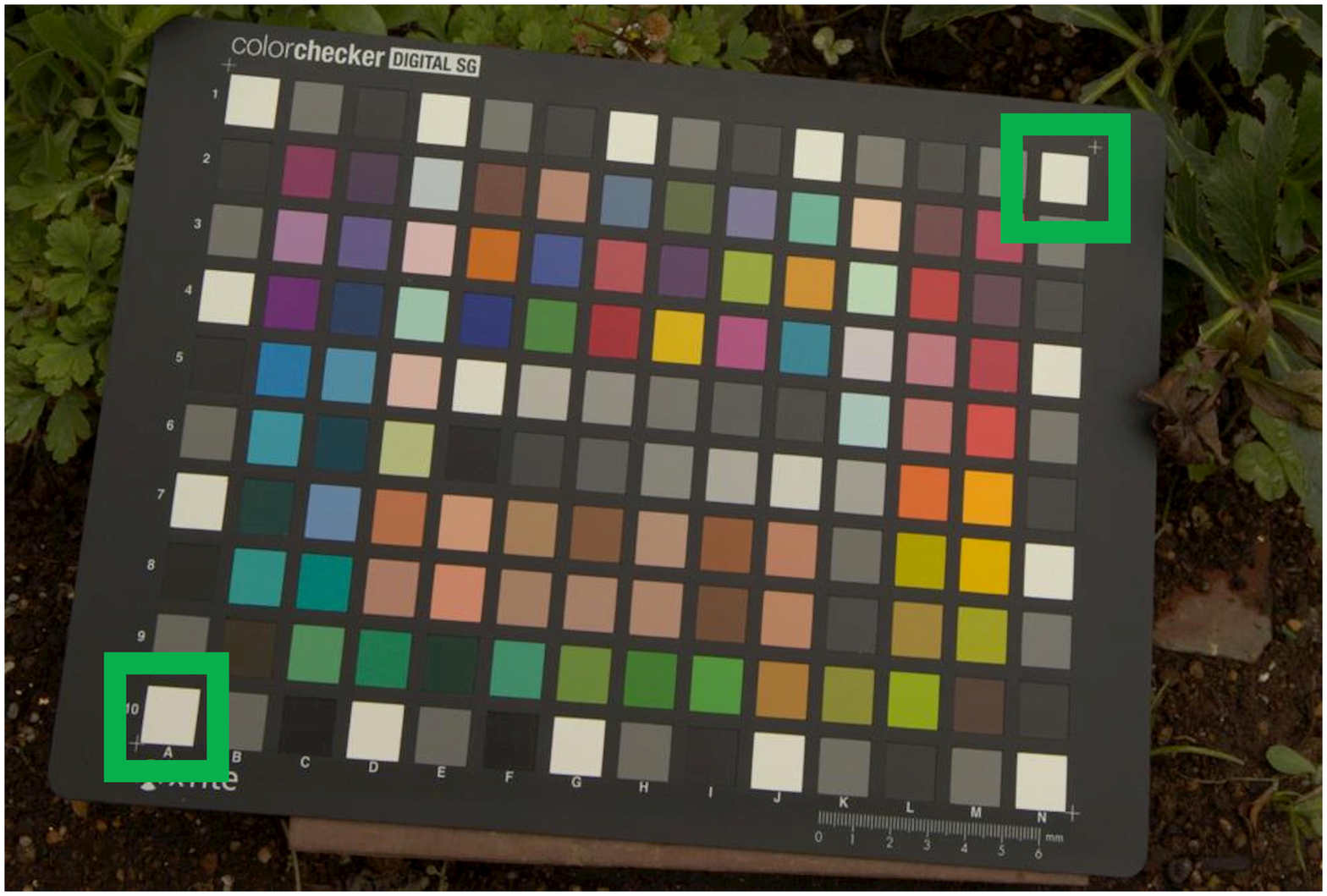}}
 \subcaption{}\label{fig:single-outside_input}\medskip
\end{minipage}
\begin{minipage}[b]{0.32\linewidth}
  \centering
  \centerline{\includegraphics[keepaspectratio, scale=0.092]{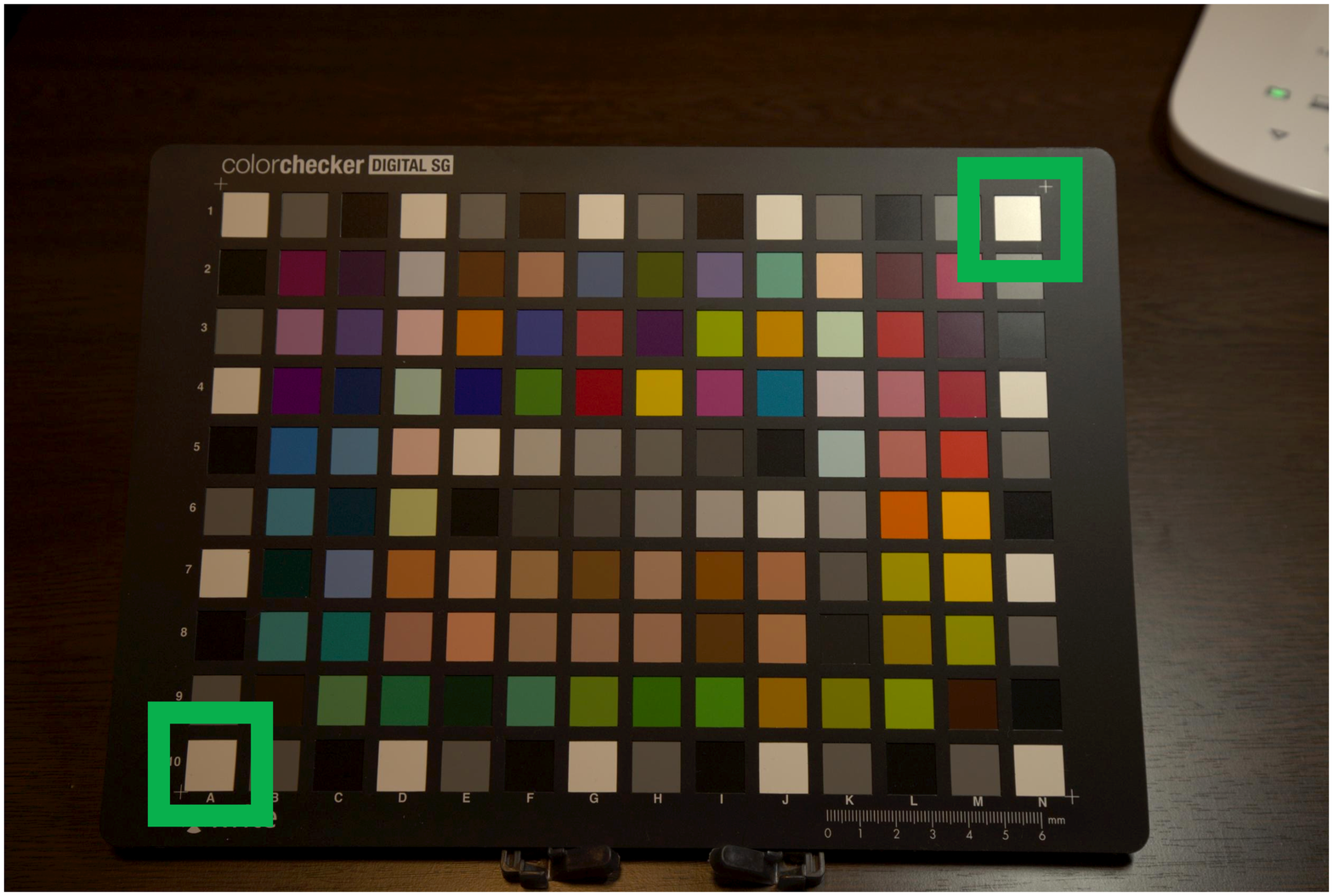}}
 \subcaption{}\label{fig:mixed_input}\medskip
\end{minipage}
\begin{minipage}[b]{0.49\linewidth}
  \centering
  \centerline{\includegraphics[keepaspectratio, scale=0.103]{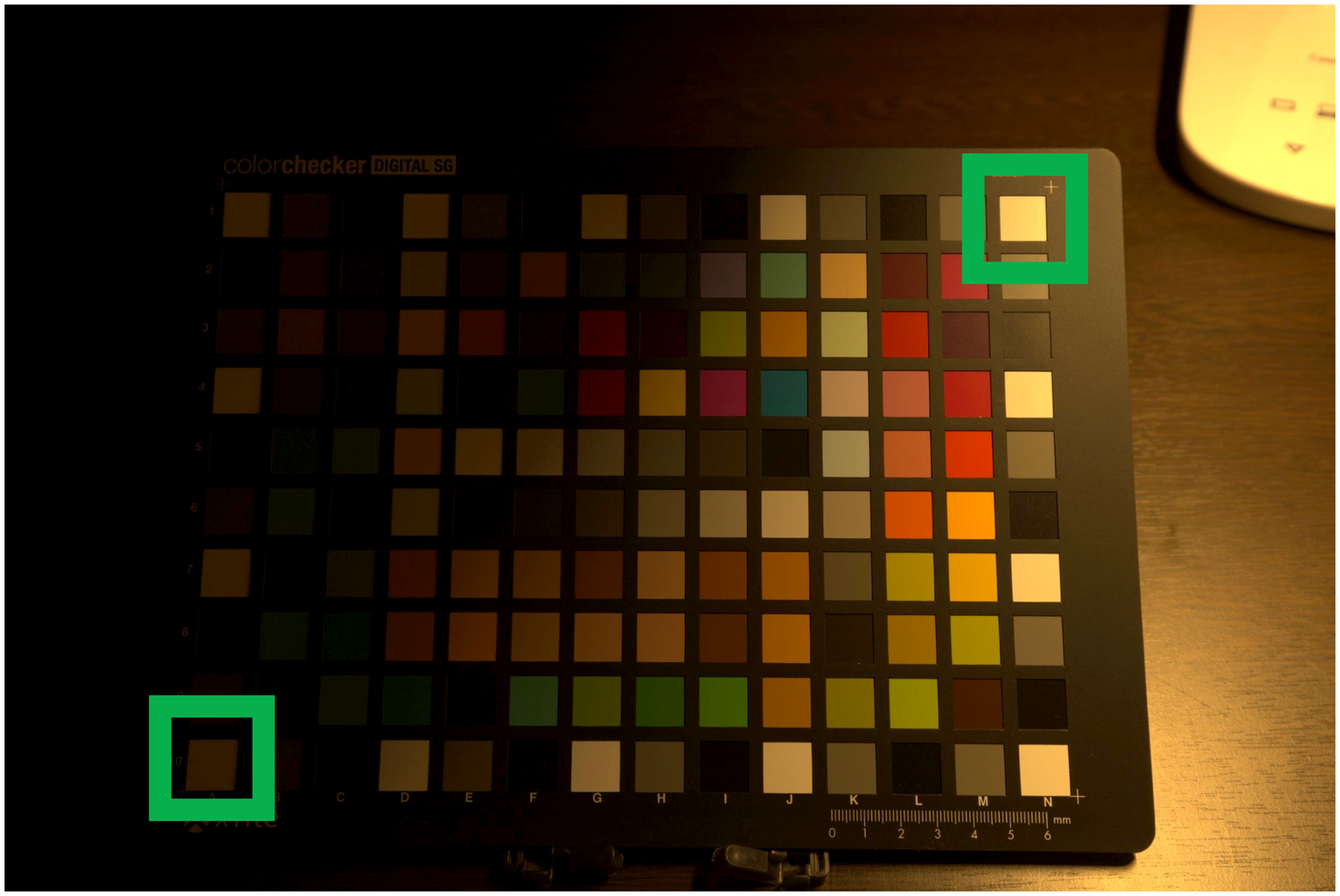}}
 \subcaption{}\label{fig:shade_input}\medskip
\end{minipage}
\begin{minipage}[b]{0.49\linewidth}
  \centering
  \centerline{\includegraphics[keepaspectratio, scale=0.107]{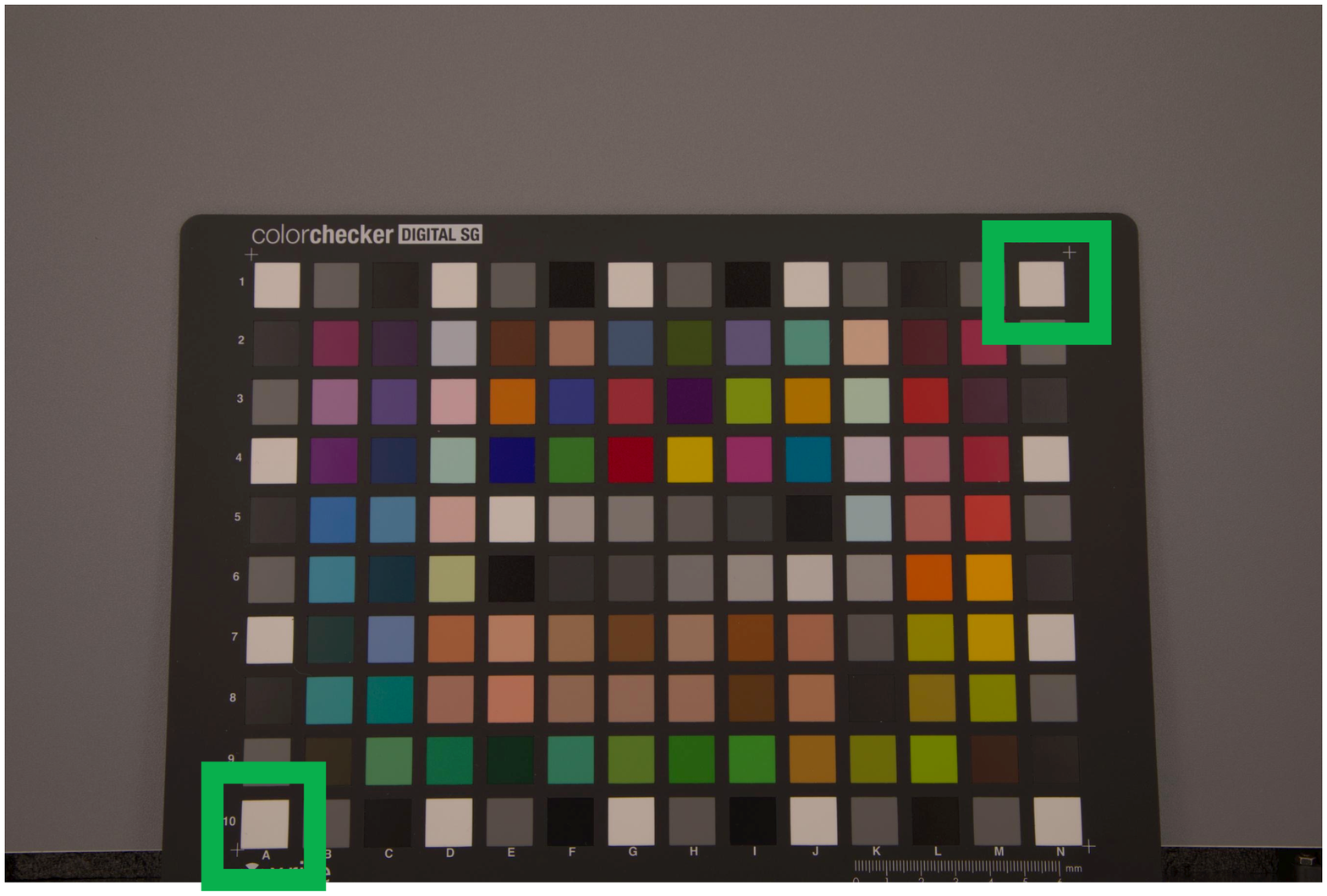}}
 \subcaption{}\label{fig:ground_truth_input}\medskip
\end{minipage}
%
\caption{Images taken under different lighting conditions.
(\subref{fig:single-inside_input}) Single artificial-illuminant, (\subref{fig:single-outside_input}) single daylight, (\subref{fig:mixed_input}) mixed illuminants, (\subref{fig:shade_input}) non-uniform illuminant, and (\subref{fig:ground_truth_input}) ground truth image.
Regions inside green squares are spatially varying white regions.}
\label{fig:Experiment_input_images}
\end{figure}

In this experiment, we used images taken under our scenario such as single, mixed, and non-uniform illuminants (see Fig. \ref{fig:Experiment_input_images}).
%
%
%
The proposed method was applied to the images in Figs. \ref{fig:Experiment_input_images} (\subref{fig:single-inside_input})--(\subref{fig:shade_input}) and compared with both the conventional white balancing and multi-color one\cite{ChengsBeyondWhite}.
Note that the proposed method and white balancing were combined with Bradford's model.
Source white points were decided from the white regions as shown in Fig. \ref{fig:Experiment_input_images} (\subref{fig:single-inside_input})--(\subref{fig:shade_input}), and the ground truth white points were selected from Fig. \ref{fig:Experiment_input_images} (\subref{fig:ground_truth_input}).
Because there were two white regions in each input image, we separately applied white balancing to the images with two source white points.
The performance of each balancing was evaluated by using the reproduction angular error\cite{ReproductionError} between a mean-pixel vector of an adjusted object's region ${\bm{P}}$ and that of the corresponding ground truth one ${\bm{Q}}$.
The reproduction error between ${\bm{P}}$ and ${\bm{Q}}$ is given by
\begin{equation}\label{reproduction_angular_error}
Err = {\frac{180}{\pi}} \: {\rm{cos}}^{-1} \left( \frac{ {{\bm{P}}} \cdot {{\bm{Q}}} }{ \|{{\bm{P}}}\| \|{{\bm{Q}}}\| } \right) \:\:\: \mathrm{[deg]}\:
.
\end{equation}
\subsection{Evaluation under single illuminants}
\begin{figure}[tb]
\captionsetup[subfigure]{justification=centering}
\begin{minipage}[b]{0.32\linewidth}
  \centering
  \centerline{\includegraphics[keepaspectratio, scale=0.3]{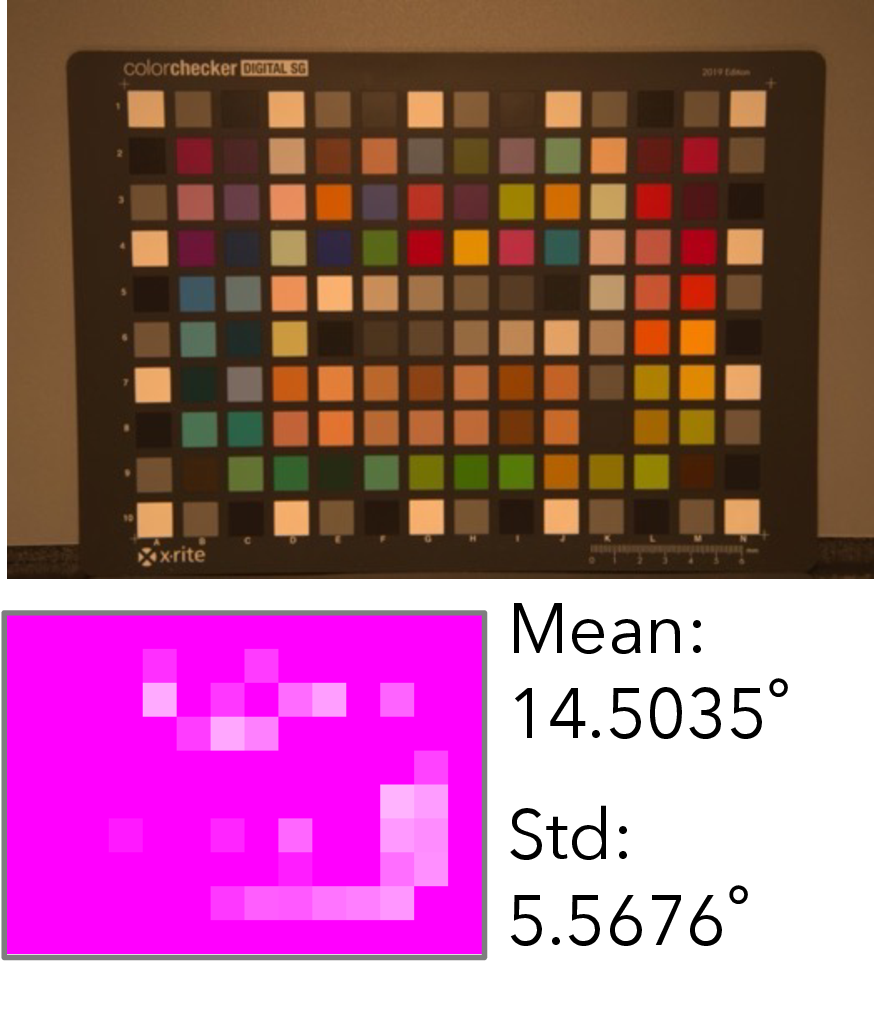}}
 \subcaption{}\label{fig:results_single-inside_input}\medskip
\end{minipage}
\begin{minipage}[b]{0.32\linewidth}
  \centering
  \centerline{\includegraphics[keepaspectratio, scale=0.3]{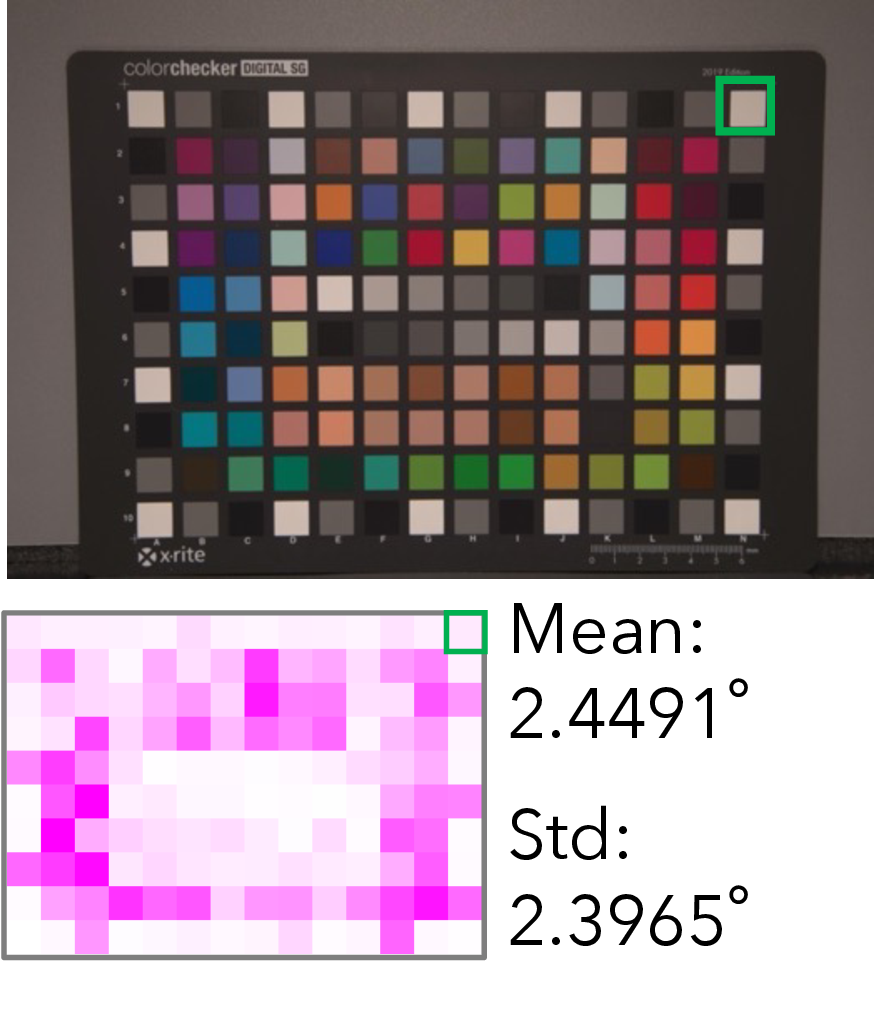}}
 \subcaption{}\label{fig:results_single-inside_WB-14}\medskip
\end{minipage}
\begin{minipage}[b]{0.32\linewidth}
  \centering
  \centerline{\includegraphics[keepaspectratio, scale=0.3]{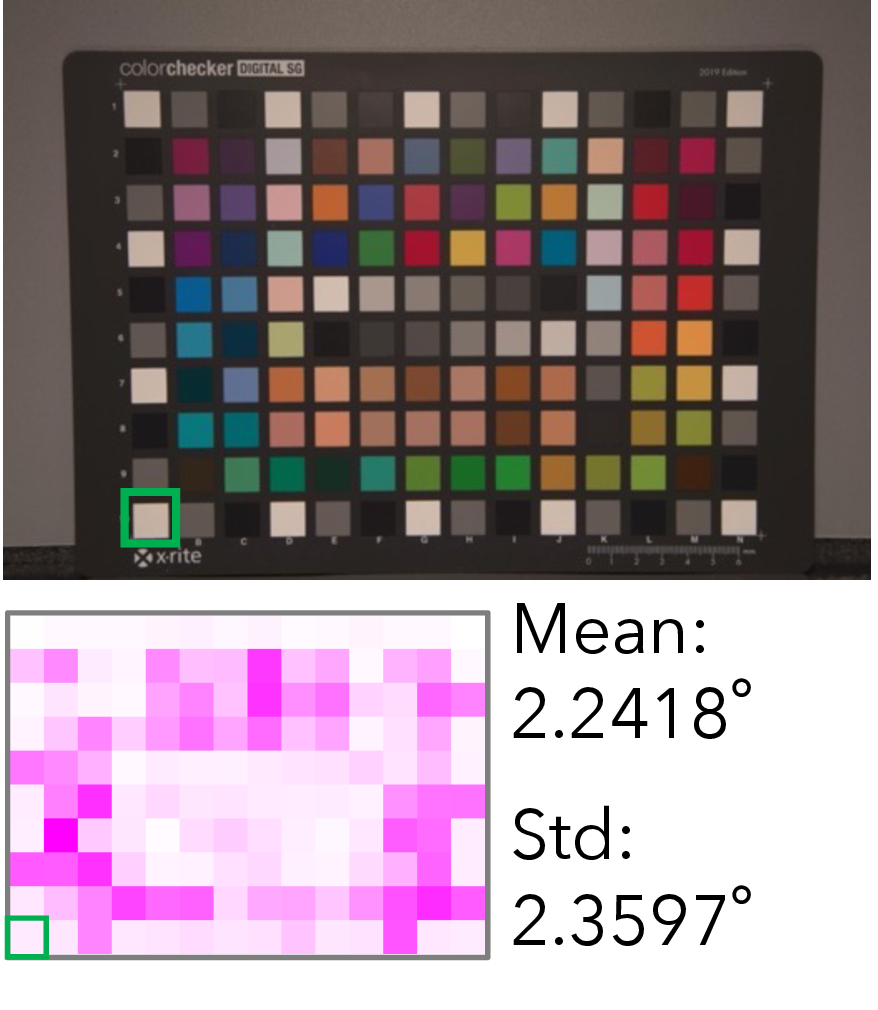}}
 \subcaption{}\label{fig:results_single-inside_WB-127}\medskip
\end{minipage}
\begin{minipage}[b]{0.4\linewidth}
  \centering
  \centerline{\includegraphics[keepaspectratio, scale=0.3]{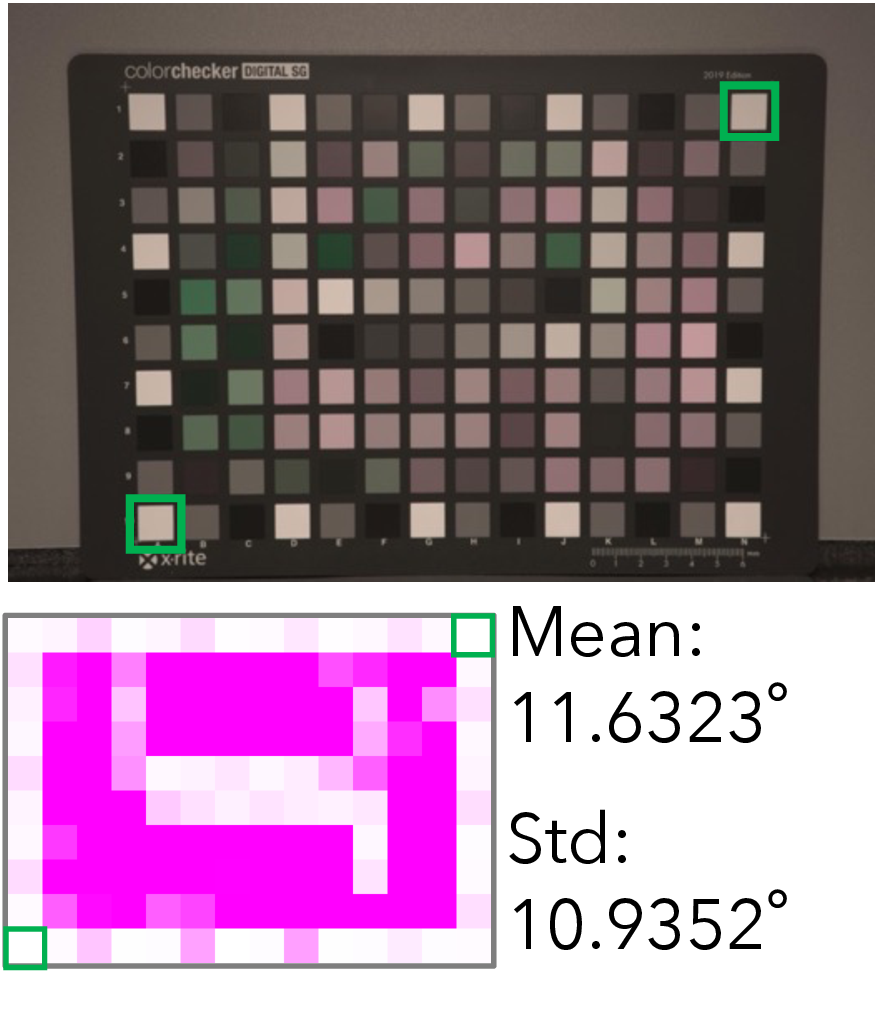}}
 \subcaption{}\label{fig:results_single-inside_Cheng}\medskip
\end{minipage}
\begin{minipage}[b]{0.4\linewidth}
  \centering
  \centerline{\includegraphics[keepaspectratio, scale=0.3]{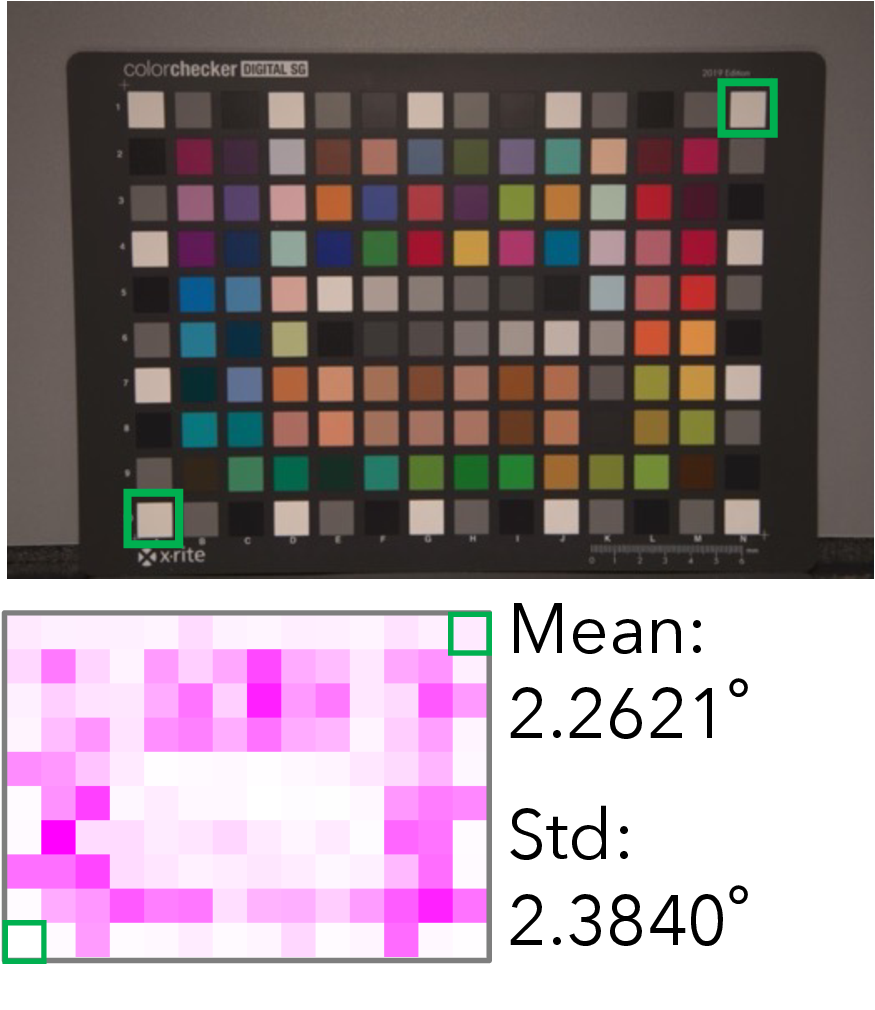}}
 \subcaption{}\label{fig:results_single-inside_n-color}\medskip
\end{minipage}
\begin{minipage}[b]{0.18\linewidth}
  \centering
  \centerline{\includegraphics[keepaspectratio, scale=0.35]{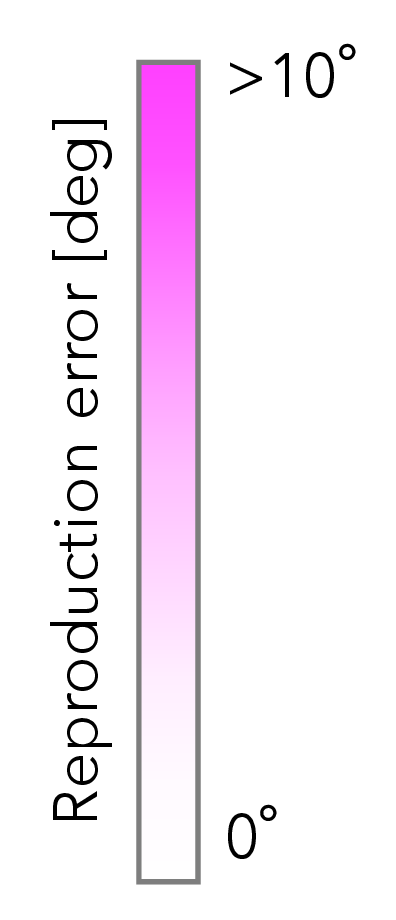}}
 \subcaption{}\label{fig:results_single-inside_color_bar}\medskip
\end{minipage}
%
\caption{Adjusted images for Fig. \ref{fig:Experiment_input_images} (\subref{fig:single-inside_input}).
(\subref{fig:results_single-inside_input}) Pre-correction, (\subref{fig:results_single-inside_WB-14}) white balancing with top-right white region, (\subref{fig:results_single-inside_WB-127}) white balancing with bottom-left white region, (\subref{fig:results_single-inside_Cheng}) conventional multi-color balancing, (\subref{fig:results_single-inside_n-color}) proposed method, and (\subref{fig:results_single-inside_color_bar}) color bar of heat maps.
}
\label{fig:Experiment_result_single-inside}
\end{figure}

\begin{figure}[tb]
\captionsetup[subfigure]{justification=centering}
\begin{minipage}[b]{0.32\linewidth}
  \centering
  \centerline{\includegraphics[keepaspectratio, scale=0.3]{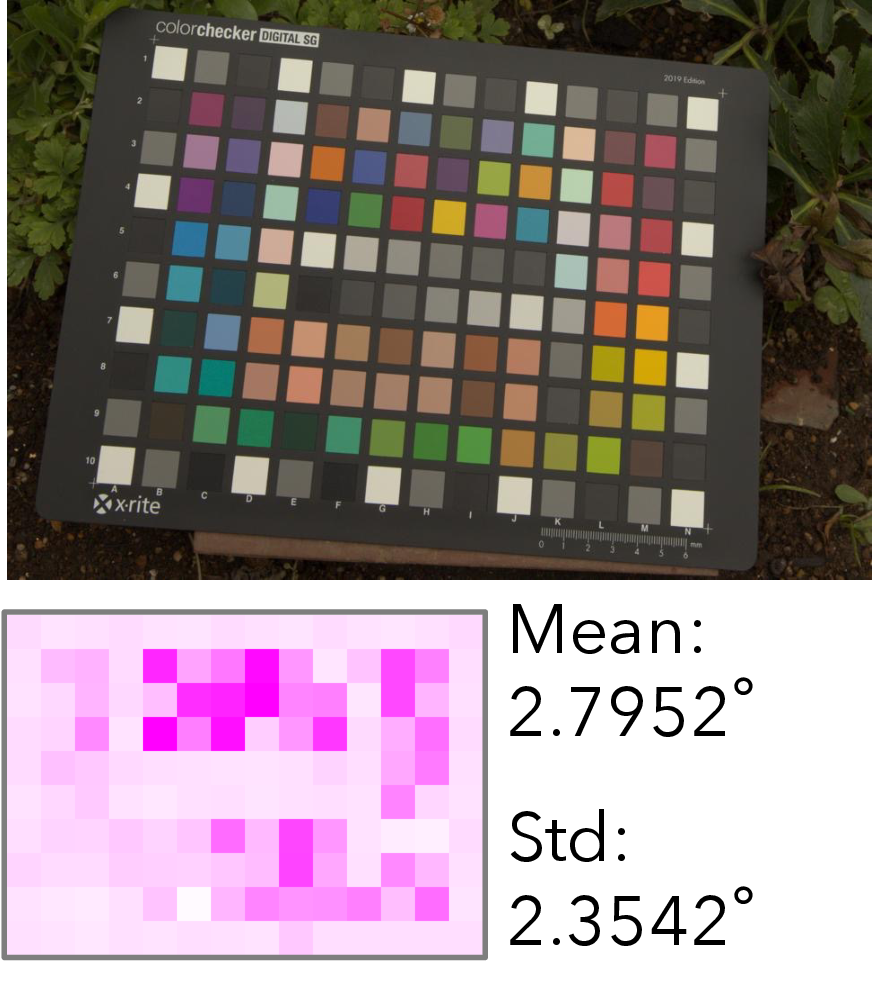}}
 \subcaption{}\label{fig:results_single-outside_input}\medskip
\end{minipage}
\begin{minipage}[b]{0.32\linewidth}
  \centering
  \centerline{\includegraphics[keepaspectratio, scale=0.3]{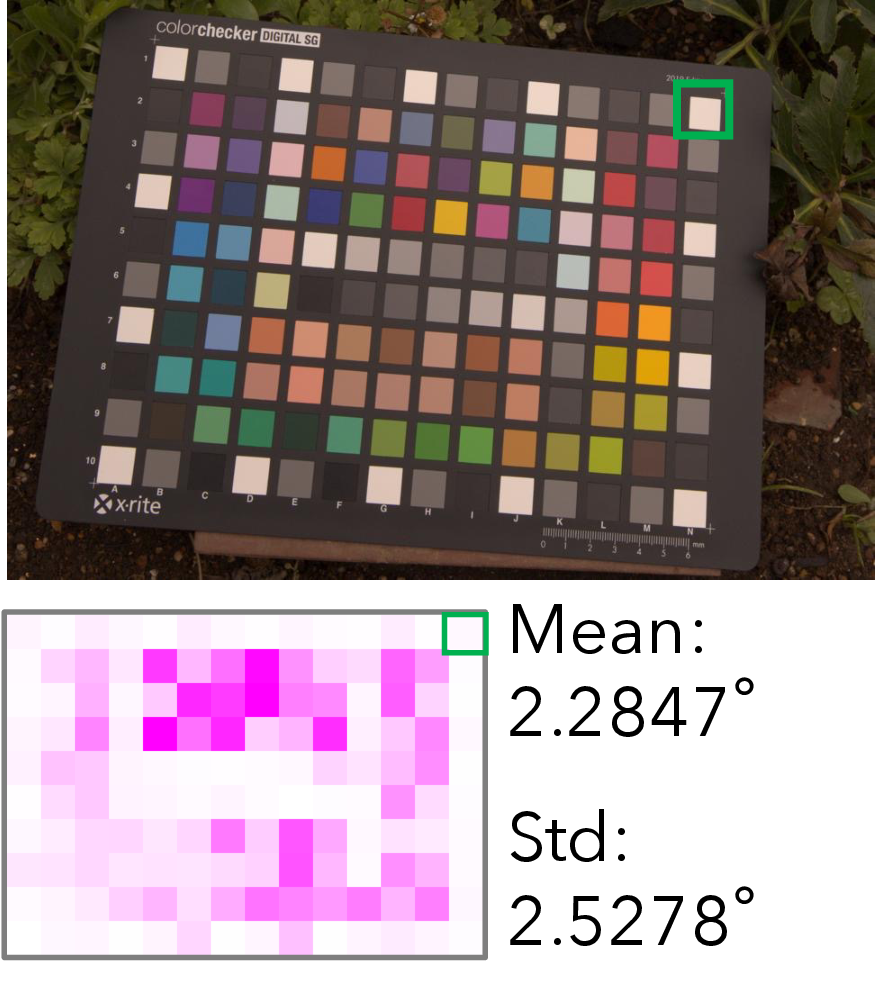}}
 \subcaption{}\label{fig:results_single-outside_WB-14}\medskip
\end{minipage}
\begin{minipage}[b]{0.32\linewidth}
  \centering
  \centerline{\includegraphics[keepaspectratio, scale=0.3]{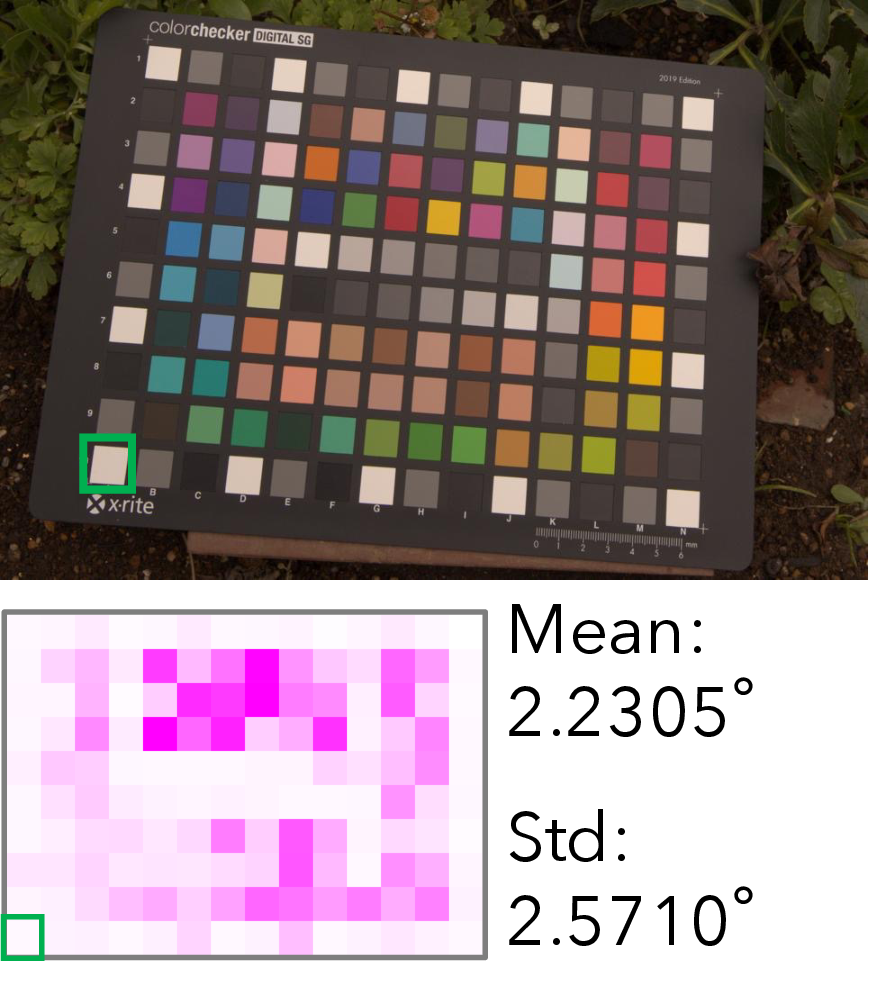}}
 \subcaption{}\label{fig:results_single-outside_WB-127}\medskip
\end{minipage}
\begin{minipage}[b]{0.4\linewidth}
  \centering
  \centerline{\includegraphics[keepaspectratio, scale=0.3]{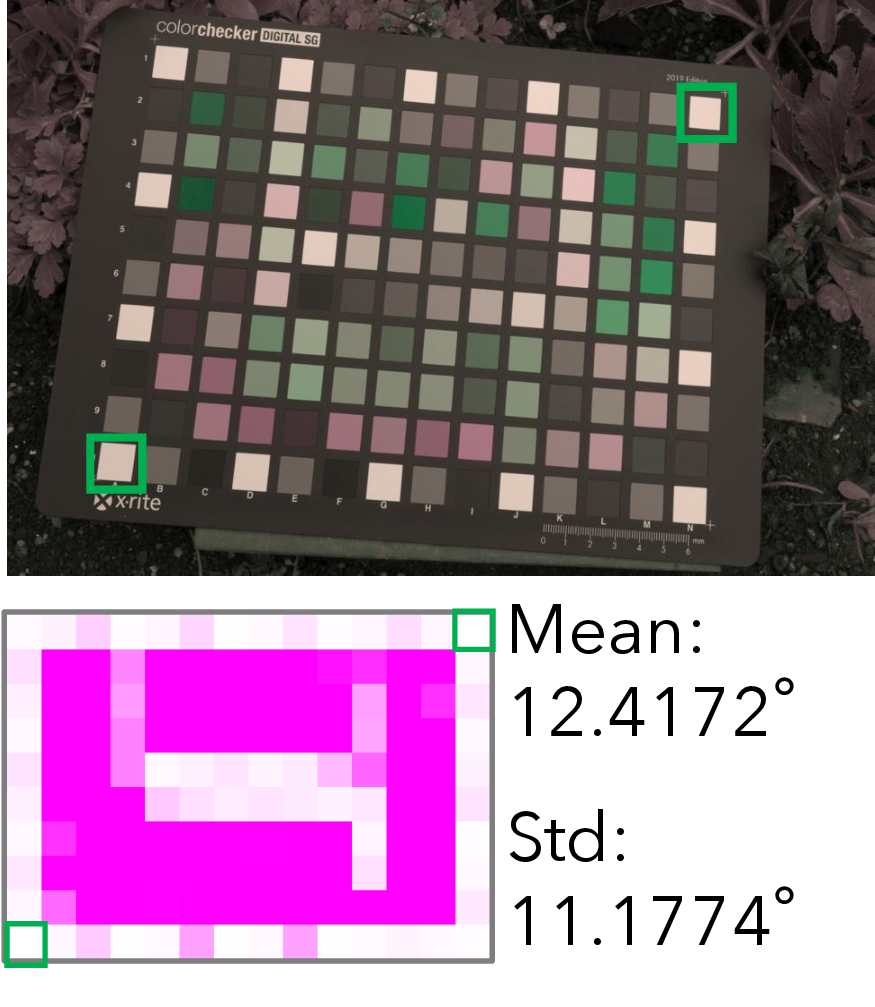}}
 \subcaption{}\label{fig:results_single-outside_Cheng}\medskip
\end{minipage}
\begin{minipage}[b]{0.4\linewidth}
  \centering
  \centerline{\includegraphics[keepaspectratio, scale=0.3]{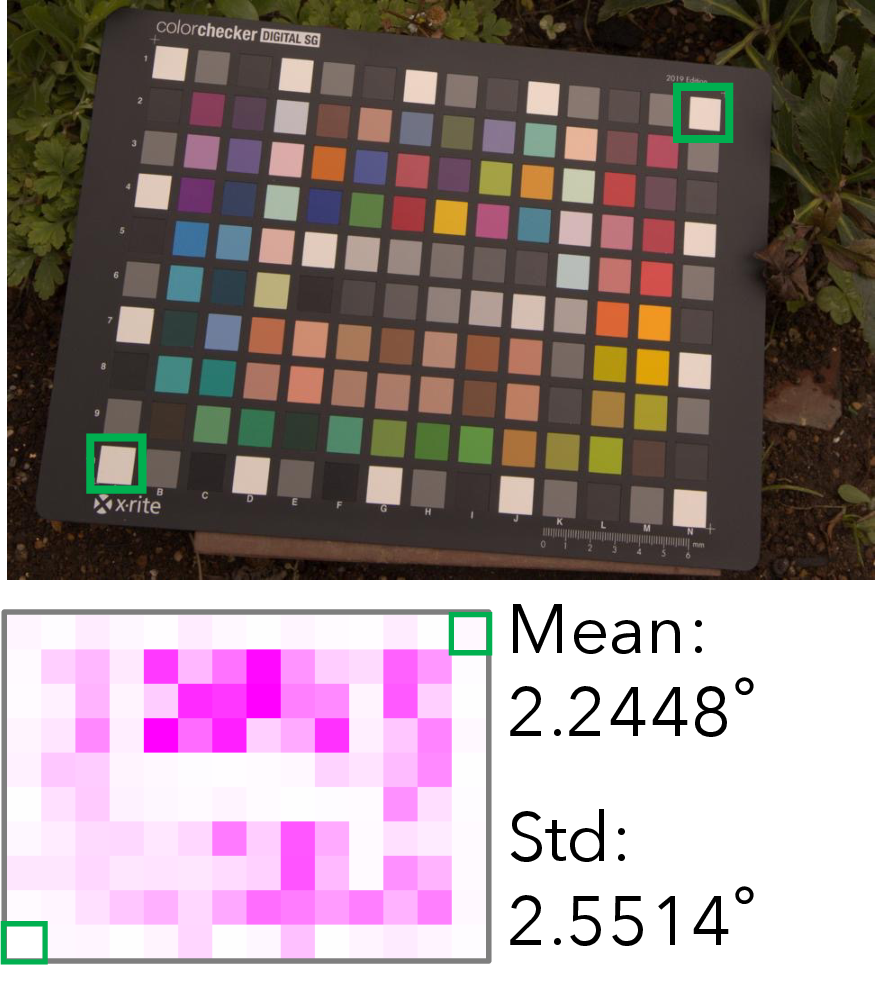}}
 \subcaption{}\label{fig:results_single-outside_n-color}\medskip
\end{minipage}
\begin{minipage}[b]{0.18\linewidth}
  \centering
  \centerline{\includegraphics[keepaspectratio, scale=0.35]{fig/Experiment/color_bar}}
 \subcaption{}\label{fig:results_single-outside_color_bar}\medskip
\end{minipage}
%
\caption{Adjusted images for Fig. \ref{fig:Experiment_input_images} (\subref{fig:single-outside_input}).
(\subref{fig:results_single-outside_input}) Pre-correction, (\subref{fig:results_single-outside_WB-14}) white balancing with top-right white region, (\subref{fig:results_single-outside_WB-127}) white balancing with bottom-left white region, (\subref{fig:results_single-outside_Cheng}) conventional multi-color balancing, (\subref{fig:results_single-outside_n-color}) proposed method, and (\subref{fig:results_single-outside_color_bar}) color bar of heat maps.
}
\label{fig:Experiment_result_single-outside}
\end{figure}

Let us discuss adjusting images taken under single illuminants.
Figs. \ref{fig:Experiment_result_single-inside} and \ref{fig:Experiment_result_single-outside} show adjusted images for Figs. \ref{fig:Experiment_input_images} (\subref{fig:single-inside_input}) and (\subref{fig:single-outside_input}).
%
%
%
%
%
%
In the figures, the heat map below each image indicates the reproduction angular error for every color patch, where ${\bm{P}}$ in (\ref{reproduction_angular_error}) is the representative pixel value of one of the color patches in an adjusted image for Figs. \ref{fig:Experiment_input_images} (\subref{fig:single-inside_input})--(\subref{fig:shade_input}), and ${\bm{Q}}$ is that for the ground truth image.
The mean value of pixel values in a color region (i.e., color patch) was used as a representative value.
Also, the mean value (Mean) and the standard variation (Std) of all color patches are shown next to each heat map.
Note that patches whose color is pure black were excluded from the calculation of Mean and Std because the metric cannot precisely measure errors against low surface reflectances.
As shown in Figs. \ref{fig:Experiment_result_single-inside} and \ref{fig:Experiment_result_single-outside}, because the single-lighting conditions were normal, conventional white balancing had a lower error than that of the input image in terms of the mean error.
Also, the proposed method performed almost the same as white balancing.
Compared with white balancing and the proposed method, conventional multi-color balancing did not reduce lighting effects on colors other than neutral colors.
This means that multi-color balancing resulted in rank deficiency under $n=2$, and this property will continuously be shown in the following experiments.
In contrast, the proposed method did not cause this problem to occur because of the use of a matrix designed from diagonal matrices.
Therefore, it is also effective even for adjusting normal illuminant conditions as well as white balancing.
\subsection{Evaluation under mixed and non-uniform illuminants}
\begin{figure}[tb]
\captionsetup[subfigure]{justification=centering}
\begin{minipage}[b]{0.32\linewidth}
  \centering
  \centerline{\includegraphics[keepaspectratio, scale=0.3]{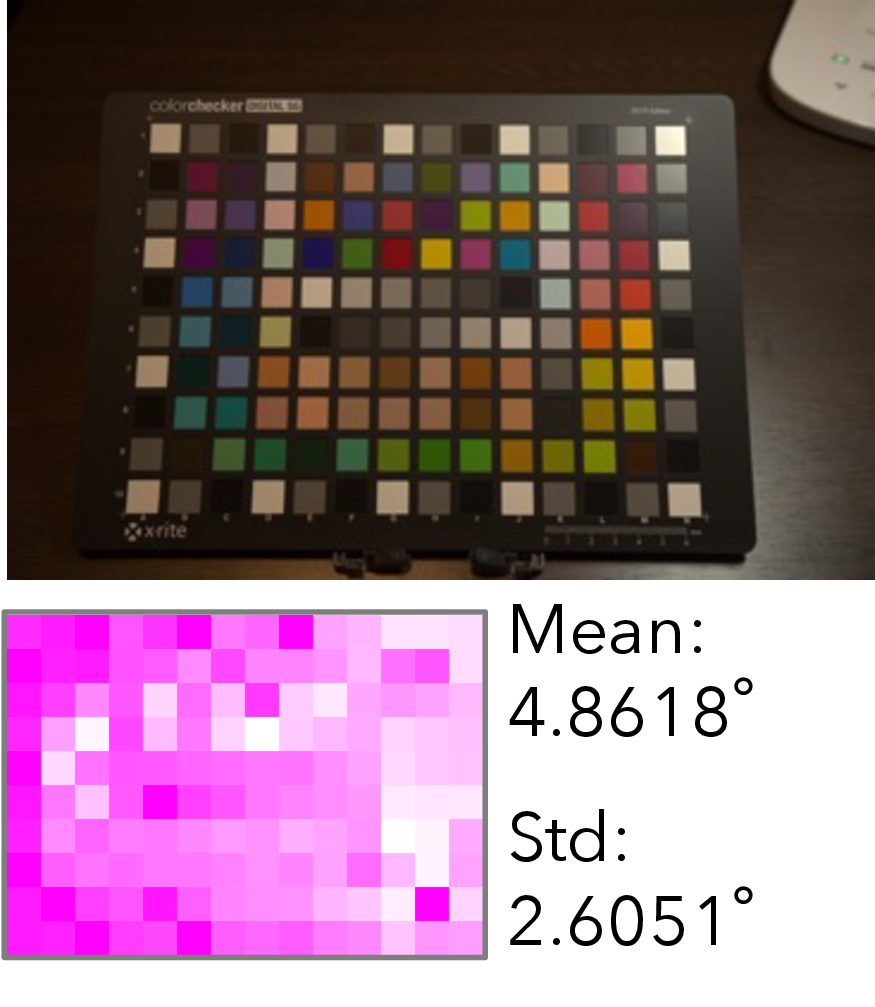}}
 \subcaption{}\label{fig:results_mixed_input}\medskip
\end{minipage}
\begin{minipage}[b]{0.32\linewidth}
  \centering
  \centerline{\includegraphics[keepaspectratio, scale=0.3]{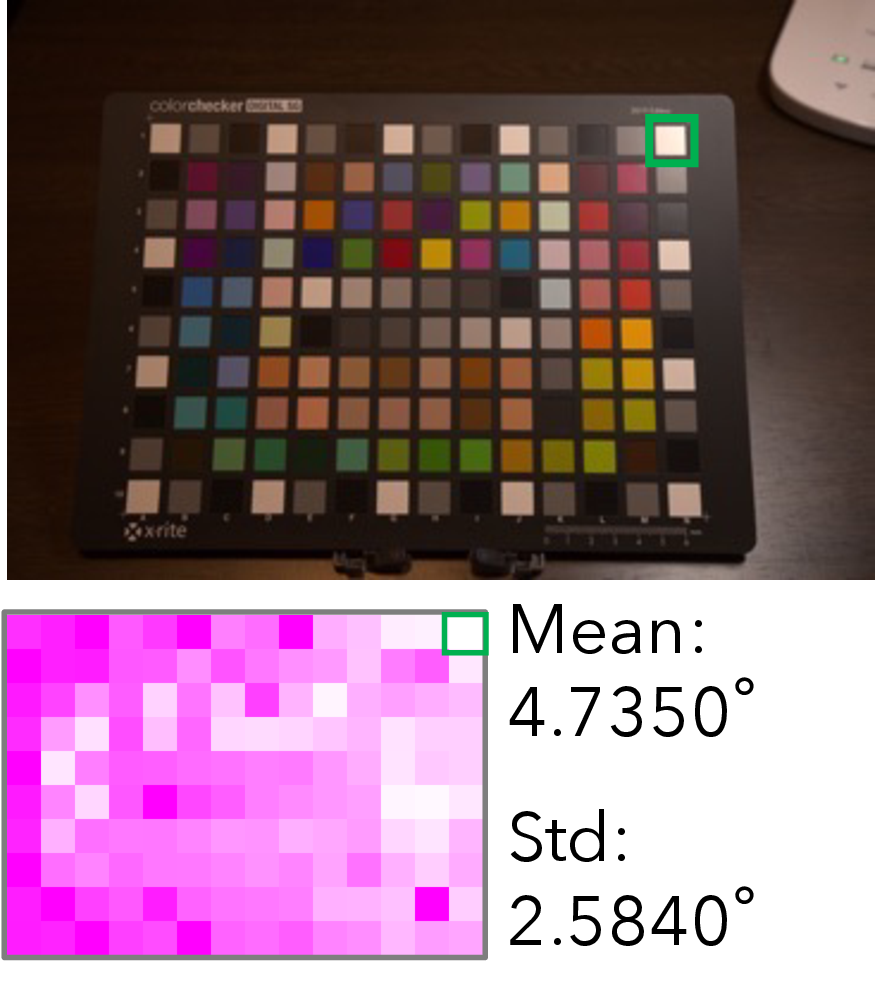}}
 \subcaption{}\label{fig:results_mixed_WB-14}\medskip
\end{minipage}
\begin{minipage}[b]{0.32\linewidth}
  \centering
  \centerline{\includegraphics[keepaspectratio, scale=0.3]{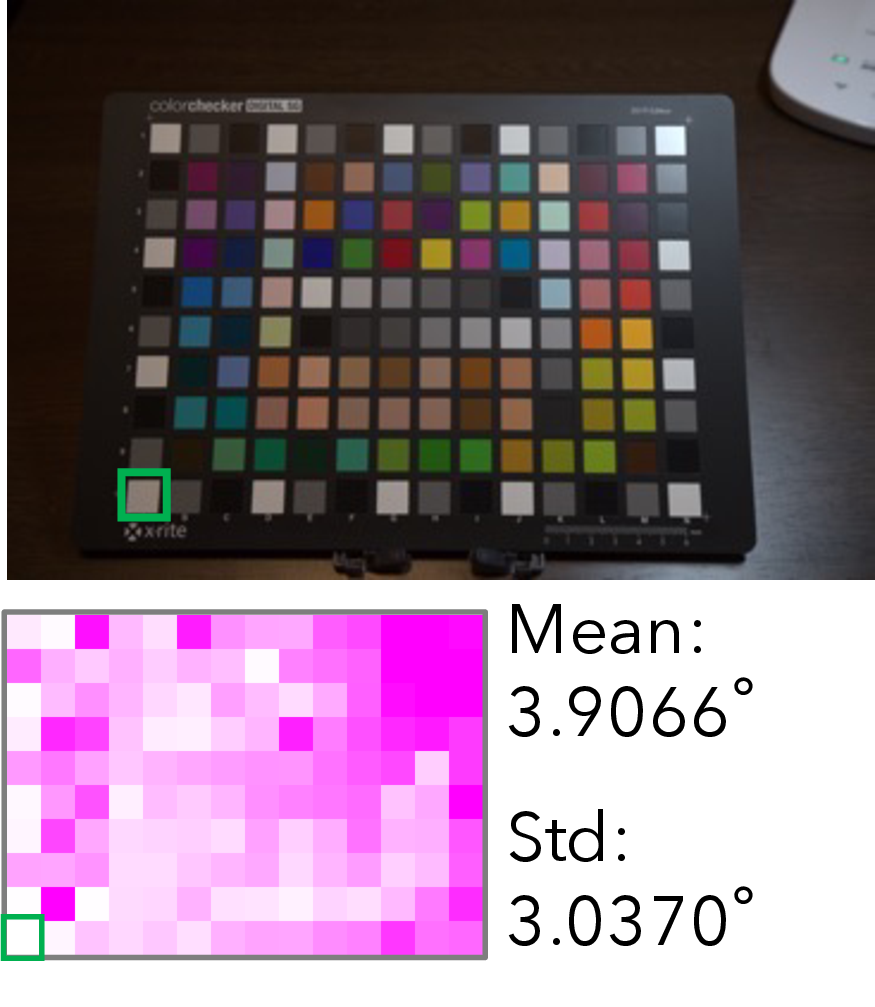}}
 \subcaption{}\label{fig:results_mixed_WB-127}\medskip
\end{minipage}
\begin{minipage}[b]{0.4\linewidth}
  \centering
  \centerline{\includegraphics[keepaspectratio, scale=0.3]{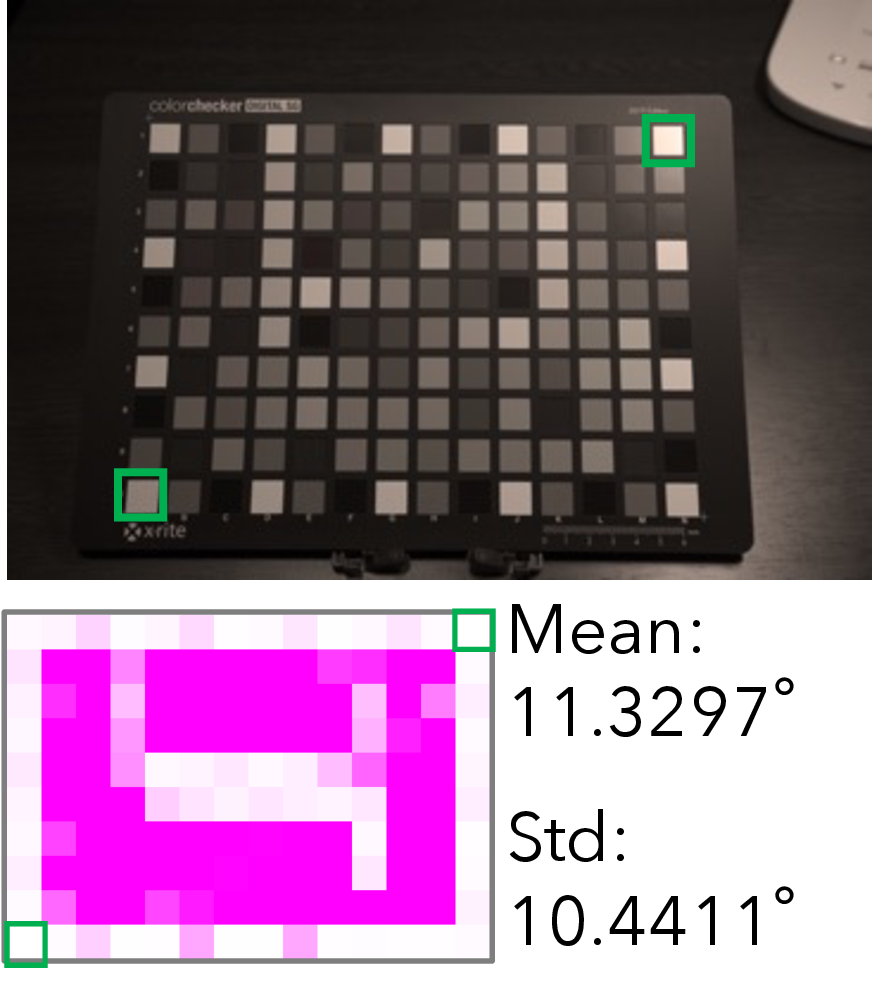}}
 \subcaption{}\label{fig:results_mixed_Cheng}\medskip
\end{minipage}
\begin{minipage}[b]{0.4\linewidth}
  \centering
  \centerline{\includegraphics[keepaspectratio, scale=0.3]{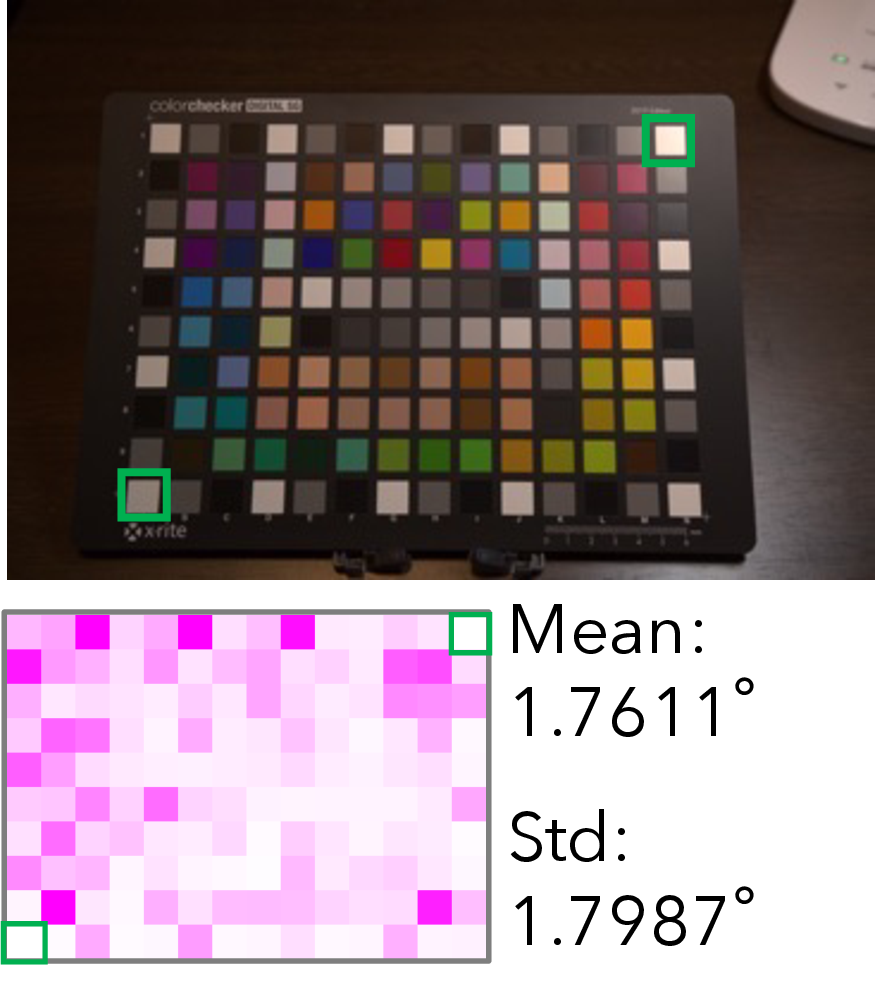}}
 \subcaption{}\label{fig:results_mixed_n-color}\medskip
\end{minipage}
\begin{minipage}[b]{0.18\linewidth}
  \centering
  \centerline{\includegraphics[keepaspectratio, scale=0.35]{fig/Experiment/color_bar}}
 \subcaption{}\label{fig:results_mixed_color_bar}\medskip
\end{minipage}
%
\caption{Adjusted images for Fig. \ref{fig:Experiment_input_images} (\subref{fig:mixed_input}).
(\subref{fig:results_mixed_input}) Pre-correction, (\subref{fig:results_mixed_WB-14}) white balancing with top-right white region, (\subref{fig:results_mixed_WB-127}) white balancing with bottom-left white region, (\subref{fig:results_mixed_Cheng}) conventional multi-color balancing, (\subref{fig:results_mixed_n-color}) proposed method, and (\subref{fig:results_mixed_color_bar}) color bar of heat maps.
}
\label{fig:Experiment_result_mixed}
\end{figure}

In this experiment, each adjustment was applied to our scenario, in particular, under mixed and non-uniform illuminants.
Fig. \ref{fig:Experiment_result_mixed} shows adjusted images for Fig. \ref{fig:Experiment_input_images} (\subref{fig:mixed_input}) taken under mixed illuminants.
%
%
%
From the figure, the proposed method exceptionally reduced lighting effects on colors under multiple illuminants in terms of mean error.
In contrast, white balancing did not adjust colors other than either of the white regions, especially for Fig. \ref{fig:Experiment_result_mixed} (\subref{fig:results_mixed_WB-14}), which had almost the same mean error as that of Fig. \ref{fig:Experiment_result_mixed} (\subref{fig:results_mixed_input}).
Thus, the proposed method can be applied to correcting images under mixed illuminants, while white balancing cannot.

\begin{figure}[tb]
\captionsetup[subfigure]{justification=centering}
\begin{minipage}[b]{0.32\linewidth}
  \centering
  \centerline{\includegraphics[keepaspectratio, scale=0.3]{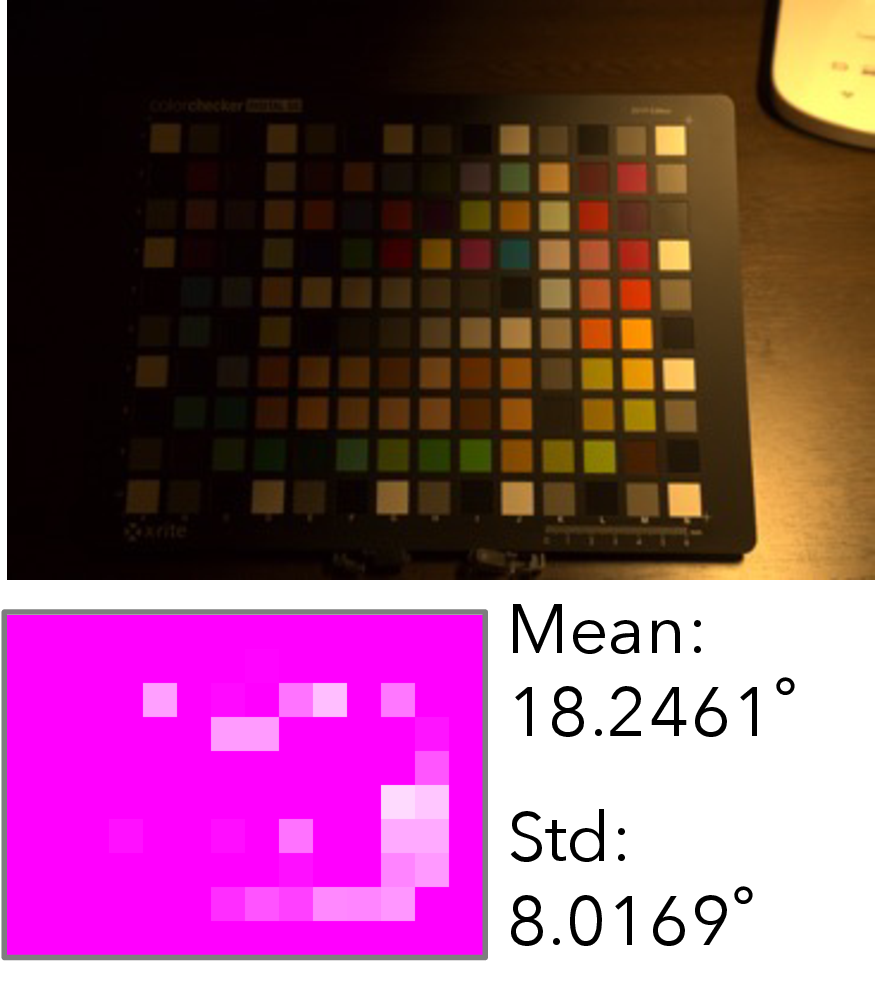}}
 \subcaption{}\label{fig:results_shade_input}\medskip
\end{minipage}
\begin{minipage}[b]{0.32\linewidth}
  \centering
  \centerline{\includegraphics[keepaspectratio, scale=0.3]{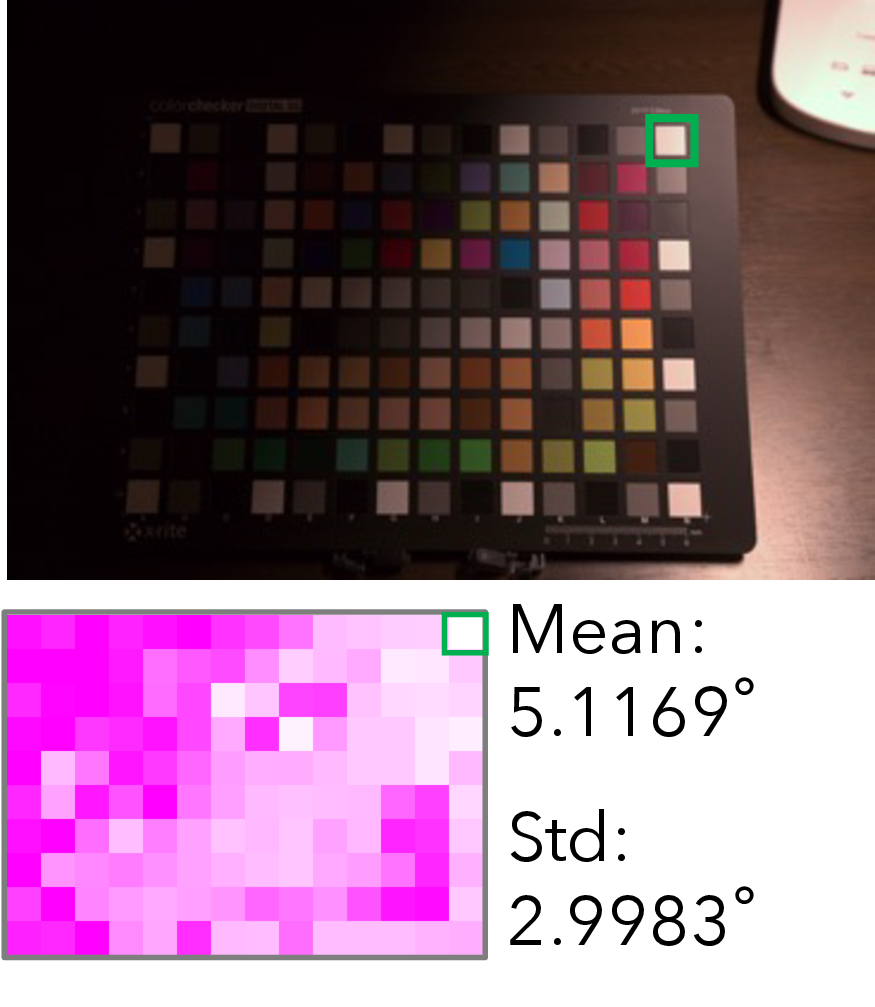}}
 \subcaption{}\label{fig:results_shade_WB-14}\medskip
\end{minipage}
\begin{minipage}[b]{0.32\linewidth}
  \centering
  \centerline{\includegraphics[keepaspectratio, scale=0.3]{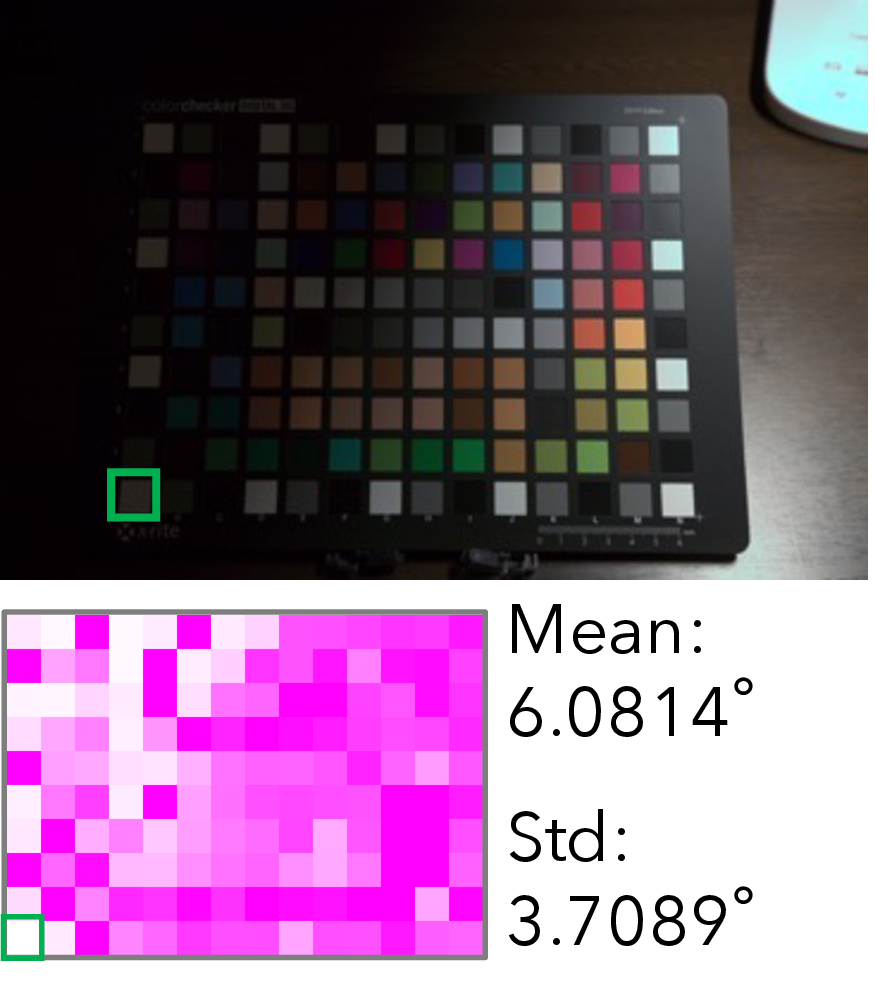}}
 \subcaption{}\label{fig:results_shade_WB-127}\medskip
\end{minipage}
\begin{minipage}[b]{0.4\linewidth}
  \centering
  \centerline{\includegraphics[keepaspectratio, scale=0.3]{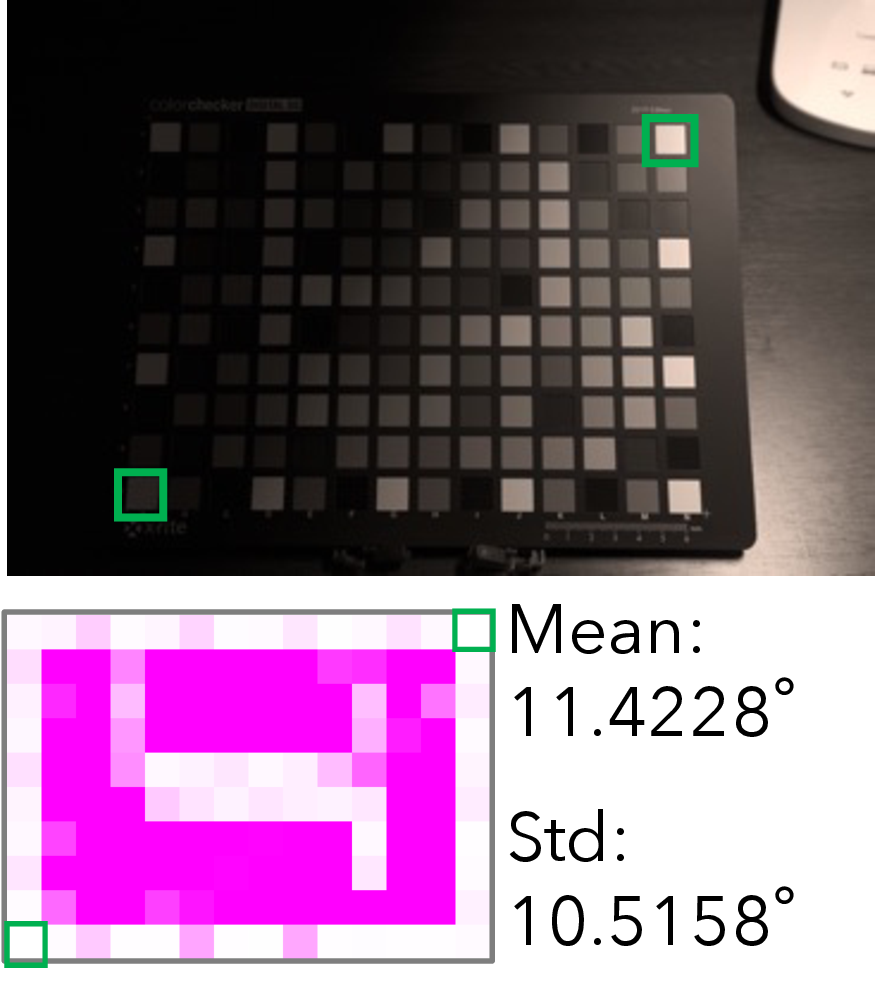}}
 \subcaption{}\label{fig:results_shade_Cheng}\medskip
\end{minipage}
\begin{minipage}[b]{0.4\linewidth}
  \centering
  \centerline{\includegraphics[keepaspectratio, scale=0.3]{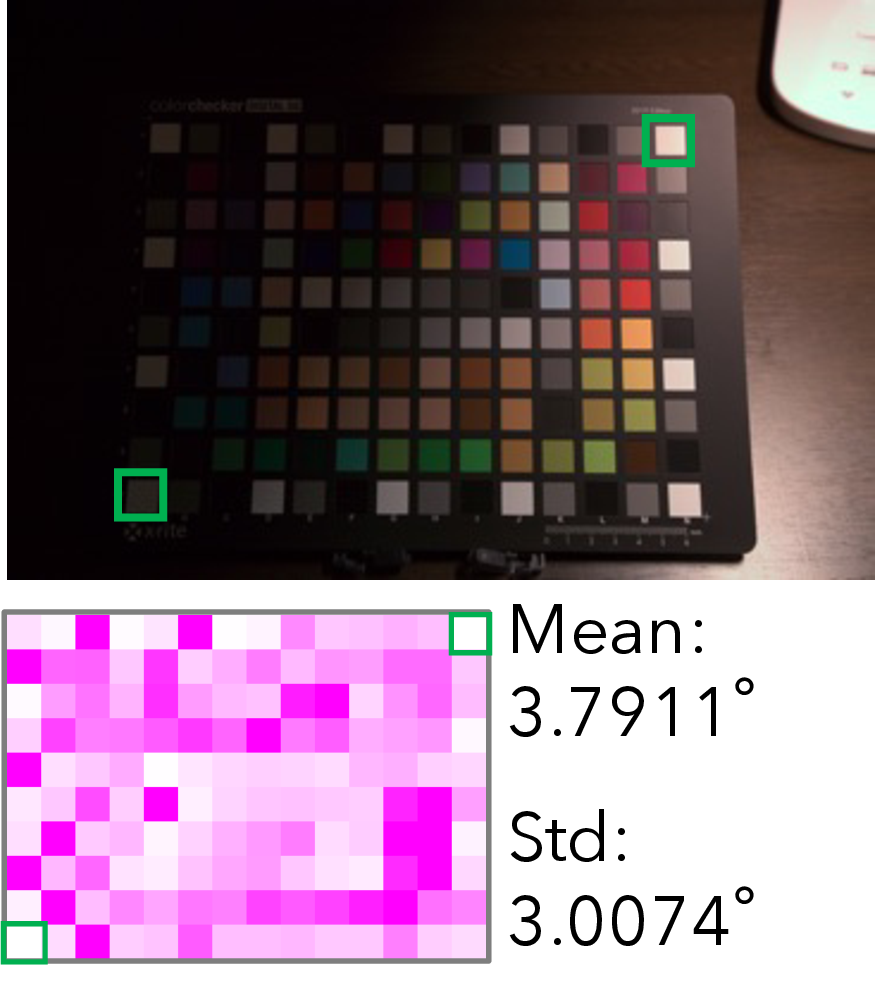}}
 \subcaption{}\label{fig:results_shade_n-color}\medskip
\end{minipage}
\begin{minipage}[b]{0.18\linewidth}
  \centering
  \centerline{\includegraphics[keepaspectratio, scale=0.35]{fig/Experiment/color_bar}}
 \subcaption{}\label{fig:results_shade_color_bar}\medskip
\end{minipage}
%
\caption{Adjusted images for Fig. \ref{fig:Experiment_input_images} (\subref{fig:shade_input}).
(\subref{fig:results_shade_input}) Pre-correction, (\subref{fig:results_shade_WB-14}) white balancing with top-right white region, (\subref{fig:results_shade_WB-127}) white balancing with bottom-left white region, (\subref{fig:results_shade_Cheng}) conventional multi-color balancing, (\subref{fig:results_shade_n-color}) proposed method, and (\subref{fig:results_shade_color_bar}) color bar of heat maps.
}
\label{fig:Experiment_result_shade}
\end{figure}

Fig. \ref{fig:Experiment_result_shade} shows adjusted images for Fig. \ref{fig:Experiment_input_images} (\subref{fig:shade_input}) covered with a large amount of shade.
%
%
%
As shown in Fig. \ref{fig:Experiment_result_shade}, under non-uniform illumination, the proposed spatially varying white balancing outperformed white balancing in terms of mean error.
In comparison with the proposed method, white balancing did not reduce lighting effects on many colors as much saturated magenta remained in the heat maps.
Therefore, the proposed method can be applied to correcting spatially varying colors under mixed and non-uniform illuminants in addition to adjusting colors under single illuminants.
\section{Conclusions}
In this paper, we proposed a novel white balancing method called ``spatially varying white balancing.''
While the conventional white balancing cannot adjust spatially varying colors under mixed or non-uniform illuminants, the proposed method enables us to correct all colors in an image under such situations.
Multiple white regions are used for designing n matrices that map spatially varying colors into corresponding ground truth colors.
Because the matrices are calculated as a diagonal matrix, the proposed method does not cause rank deficiency, unlike conventional multi-color balancing.
In an experiment, the proposed method was demonstrated to outperform the conventional white and multi-color balance adjustments under single, mixed, and non-uniform illuminants.

\bibliographystyle{ieeetr}
\bibliography{references}

\end{document}